\documentclass[conference]{IEEEtran}
\IEEEoverridecommandlockouts
\usepackage[noadjust]{cite}
\usepackage{amsmath,amssymb,amsfonts}
\usepackage{algorithmic}
\usepackage{graphicx}
\usepackage{textcomp}
\usepackage{xcolor}
\usepackage{multirow}
\usepackage[T1]{fontenc}  
\usepackage[utf8]{inputenc}  
\usepackage{authblk}  
\usepackage{stfloats}
\usepackage{float}
\usepackage[section]{placeins}
\usepackage{hyperref}
\usepackage[hyphenbreaks]{breakurl}

\def\BibTeX{{\rm B\kern-.05em{\sc i\kern-.025em b}\kern-.08em
		T\kern-.1667em\lower.7ex\hbox{E}\kern-.125emX}}

\begin{document}
	
	\title{Resisting Out-of-Distribution Data Problem in Perturbation of XAI}
	
	\author[1]{Luyu QIU}  
	\author[1]{Yi YANG}  
	\author[1]{Caleb Chen CAO}  
	\author[2]{Jing LIU}  
	\author[3]{Yueyuan ZHENG}  
	\author[3]{Hilary Hei Ting NGAI}  
	\author[3]{\\Janet HSIAO} 
	\author[1]{Lei CHEN} 
	\affil[1]{Huawei Research Hong Kong}  
	\affil[2]{Huawei Technologies} 
	\affil[3]{Department of Psychology, University of Hong Kong} 
	
	\maketitle
	
	\begin{abstract}
		With the rapid development of eXplainable Artificial Intelligence (XAI), perturbation-based XAI algorithms have become quite popular due to their effectiveness and ease of implementation. 
		The vast majority of perturbation-based XAI techniques face the challenge of Out-of-Distribution (OoD) data -- an artifact of randomly perturbed data becoming inconsistent with the original dataset. OoD data leads to the over-confidence problem in model predictions, making the existing XAI approaches unreliable.
		To our best knowledge, OoD data problem in perturbation-based XAI algorithms has not been adequately addressed in the literature. 
		In this work, we address this OoD data problem by designing an additional module quantifying the affinity between the perturbed data and the original dataset distribution, which is integrated into the process of aggregation.
		Our solution is shown to be compatible with the most popular perturbation-based XAI algorithms, such as RISE, OCCLUSION, and LIME.  
		Experiments have confirmed that our methods demonstrate a significant improvement on general cases using both computational and cognitive metrics. 
		Especially in the case of degradation, our proposed approach demonstrates outstanding performance comparing to baselines. 
		Besides, our solution also resolves a fundamental problem with the faithfulness indicator, a commonly used evaluation metric of XAI algorithms that appears to be sensitive to the OoD issue.
	\end{abstract}
	
	
	\section{Introduction}\label{sec:intro}
	
	Artificial Intelligence (AI) technologies have been applied in various domains, manifesting their extraordinary power to the world. 
	However, explainablity is becoming arguably one of the most crucial challenges.
	Business owners are reluctant to rely on the black-box AI models due to the lack of explainability and transparency, which hinders the development and applications of this formidable technology. 
	To enable AI to be truly integrated into the business processes and be accepted by the wider community, eXplainable Artificial Intelligence (XAI) has been initiated \cite{XAI_survey1}. 
	Moreover, the General Data Protection Regulation (GDPR), proposed by the European Parliament, stipulates that every individual has the right to obtain a `meaningful explanation of the logic involved' in personal data processing \cite{XAI_survey2}. 
	Consequently, XAI approaches, regarded as an essential condition for AI applications, have attracted an increasing number of research interests \cite{taxonomy1, XAI_survey1, taxonomy3, taxonomy4, taxonomy5, taxonomy6, taxonomy7}. 
	
	Methodological speaking, XAI methods can be divided into two groups - gradient-based and perturbation-based. 
	Gradient-based methods compute the derivative of neural networks through the back-propagation procedure and provide visualization of active neurons' importance in each layer \cite{gradcam}. Gradient-based methods are based on the gradient information of the model, which limits their applicability to neural networks mostly. 
	Comparatively, perturbation-based algorithms focus on the relationship between different perturbed inputs and the models' outputs, which are model-agnosic and easily adopted in many real-world applications \cite{aix360-sept-2019, captum}. 
	Typically, a perturbation-based algorithm edits or occludes a given sample, and generates multiple modified samples, which are further processed for interpretation purpose \cite{perturbation-based_xai, fong2017interpretable}. There are well-known XAI algorithms such as RISE \cite{petsiuk2018rise}, OCCLUSION \cite{zeiler2014visualizing}, LIME \cite{lime}, and Extremal perturbation \cite{extremal}. 
	Nevertheless, perturbation-based XAI methods still face multiple challenges, one of which is the Out-of-Distribution (OoD) data problem \cite{OOD_inpainting}. 
	
	\textbf{Challenge of OoD data}: OoD data indicate samples that are inconsistent with the distribution of the original training dataset. In contrast, In-Distribution(InD) data follow the original data distribution. Studies \cite{nguyen2015deep, OoD_detect_baseline, szegedy2014intriguing} give an astonishing conclusion that the MNIST image classifier may produce a prediction with the 91\% confidence for an input which is a "Gaussian noise" image. This unreasonable result is because the AI model produces spurious predictions on the OoD Gaussian noise image. Such artifacts inevitably result in the AI model producing spurious predictions.  
	
	\textbf{OoD data in perturbation-based XAI}:
	Similarly, OoD data problem exists in XAI methods.
	Random perturbations may generate both InD and OoD samples.	These OoD samples are usually served to generate surrogate models, which makes the XAI algorithms unreliable. 
	Recent studies \cite{chang2018explaining, OOD_inpainting} attempted to address this issue by creating perturbed samples which are more visually plausible by in-painting masked out regions with generative models. 
	However, generative models do not necessarily produce InD samples, therefore fail to eliminate the OoD data problem fully. Additionally, the computation of generative models is highly time-consuming.  
	Another intuitive solution of OoD data problem is to simply detect and drop those OoD samples and obtain explanations by InD data only - one can evaluate the degree of OoD for each sample of the give dataset, and manually determine a threshold indicating a sample is OoD or not. However, such approach fails to discriminate the samples that are above the threshold. Moreover, it may be too difficult to fine-tune the threshold for end-users. Since every perturbed sample has a different degree of InD, a better solution is to quantify the OoD extent of each sample and integrate it into the explanation process.
	
	In this paper, we propose a general solution to address this OoD data problem by handling the shortcomings of the previous approaches. 
	Specifically, we design an OoD block that calculates an `Inlier Score', approximating the probability that a sample is generated from the original dataset distribution. 
	OoD block is designed based on anomaly detection techniques and produces inlier scores, which are then used in the process of explanation generation to reduce the prediction over-confidence induced by the OoD data problem.
	We apply this approach to three popular perturbation-based XAI methods known as RISE, OCCLUSION and LIME. By adopting our proposal, the improved versions of these methods are referred as  $RISE^+$, $OCCLUSION^+$, and $LIME^+$ respectively.

	With the same rationale, we point out that OoD data problem causes a fundamental bias in the calculation of faithfulness \cite{petsiuk2018rise}, a popular quality indicator in XAI. 
	We apply our solution to ameliorate this metric and propose the an improved indicator for faithfulness.
	
	We summarize our contributions of this paper as follows:
	\begin{enumerate}
		\item An improved general framework of perturbation-based XAI methods with an OoD block that produces `Inlier Scores' to adjust over-confident predictions;
		\item Illustration of how to improve three popular methods, namely RISE, OCCLUSION and LIME, with our framework and comprehensive experimental analysis of the effectiveness of our proposed methods $RISE^+$, $OCCLUSION^+$ and $LIME^+$.
		\item Revelation of the fundamental flaw inherent to the faithfulness indicator and its correction. 
		\item Application of cognitive metric’s framework on XAI evaluation and the design of user study.
	\end{enumerate} 
	
	\section{Related work}\label{sec:related_work}
	\subsection{Perturbation-based XAI methods}
	Perturbation-based methods are independent of the AI model structure, as they examine the relationship between inputs and outputs. 
	These methods typically perturb (i.e., remove) an input region and use the resultant classification probability change as the attribution value for that region. 
	Here are several perturbation-based XAI methods that output is saliency maps. 
	RISE \cite{petsiuk2018rise} occludes an image using random patterns and weighs the changes in the output by the perturbing masks. 
	RISE outperforms existing explanation approaches in terms of automatic causal metrics and performs competitively in terms of the human-centric metric. 
	OCCLUSION \cite{zeiler2014visualizing} uses a sliding patch occlude part of the input and records change in output. 
	LIME \cite{lime} learns a local model based on the perturbed data to get the feature importance. 
	These three algorithms belong to the class of attribution methods and are widely used in computer-vision scenarios for XAI. 
	Attribution approaches explain classifiers by generating saliency maps that show the importance of each pixel in the model’s decision. 
	In deep learning, \cite{simonyan2013deep} first introduces saliency map as a visualization of the gradient of the network’s output with respect to the input image, and the attribution method has proliferated in recent years. 
	
	Despite the popularity, extensive exploration of the attribution algorithms have also revealed their substantial limitations \cite{rudin2019stop, ghorbani2019interpretation}. 
	For example, \cite{slack2020fooling} found that if there are differences in distribution between perturbation samples and input data, the explanation methods such as LIME \cite{lime} and SHAP \cite{SHAP} can be misled to generate innocuous explanations for biased classifiers. 
	The existence of out-of-distribution samples is an inevitable problem since current perturbing techniques simply grey out \cite{zeiler2014visualizing, lime}, blur \cite{fong2017interpretable, extremal} or add noises to the input data. 
	The existing attempts to tackle the OoD data problem, such as generative models that in-paint the masked out regions to synthesize realistic data \cite{chang2018explaining, OOD_inpainting}, cannot guarantee that the generated images still belong to the true distribution, even though the generated samples are visually plausible. 
	Moreover, it is questionable whether this approach can produce a reasonable explanation as the features are replaced with synthesis conditioned on in-painting algorithms rather than directly removed.
	
	To the best of our knowledge, our work is the first attempt to tackle the OoD data problem based on anomaly detection techniques. 
	In the following section, we carry out a comprehensive study of anomaly detection methodologies to design our solution.

	\subsection{Anomaly detection techniques}
	Anomaly detection techniques can be classified into three categories: supervised, semi-supervised and unsupervised depending on the availability of OoD data labels. 
	We discuss each of these categories below.
	
	\subsubsection{Supervised approaches} The labels of InD and OoD samples are available to distinguish the difference between them in supervised approaches. 
	In \cite{AnomalyDetection1}, a neural network uncertainty quantization method based on negative log-likelihood gradient information is proposed.
	An anomaly detector can be trained according to the uncertainty measure of some OoD data. 
	Study \cite{AnomalyDetection2} estimates the probability density of the test samples on DNN feature spaces. 
	Specifically, class-conditional Gaussian distributions are fitted to pre-trained features first. 
	Then, a confidence score based on the Mahalanobis distance with respect to the closest class conditional distribution is calculated. 
	Its parameters were chosen using the empirical means and the corresponding empirical covariance matrix calculated for the training sample. 
	To further improve the performance, confidence scores from different layers of DNN were combined by using weighted averaging.
	The weight of each layer was learned by training a logistic regression detector on labeled samples containing both InD and OoD data. 
	The method was shown to be robust to OoD examples. 
	In \cite{AnomalyDetection3}, the authors trained a detector on representations derived from responses generated from a set of classifiers trained by applying various natural transformations for InD and OoD training data. 
	In \cite{AnomalyDetection4}, authors have proposed a trust score to estimate whether the prediction of test examples by a classifier can be trusted. 
	The idea is that if the classifier predicts a label that is farther than the closest label, then it may be an unreliable example.
	
	\subsubsection{Semi-supervised approaches} Anomaly detection techniques are referred to as semi-supervised if unlabeled contaminated data is utilized in addition to labeled information of InD class. 
	They are more widely adopted comparing to supervised ones because these techniques avoid knowing whether the unlabeled instance is InD or OoD. 
	In \cite{AnomalyDetection5}, the authors presented a likelihood ratio-based method that empowered deep generative models to differentiate between InD and OoD.
	
	\subsubsection{Unsupervised approaches} Unsupervised anomaly detection approaches have an advantage because they do not need labeled data. 
	The rationale behind \cite{OoD_detect_baseline} is that well-trained neural networks tend to assign a higher predicted probability to InD samples than to OoD samples. 
	So a threshold is set to compare with predicted probability and then OoD examples can be detected. 
	The approach is no longer applicable if the classifier does not separate the maximum value of the predicted distribution fully enough with respect to InD and OoD examples. 
	In \cite{AnomalyDetection7}, authors observed that using temperature scaling and adding small perturbations to the input can well separate the softmax score distribution between InD and OoD images. 
	GAN-based architecture is applied in the reconstruction error-based OoD detection approach in \cite{AnomalyDetection8}. 
	In \cite{AnomalyDetection9}, a degenerated prior network architecture is proposed that can separate model-level uncertainty from data-level uncertainty by the way of prior entropy. 
	Meanwhile, the authors addressed a concentration perturbation algorithm, which adds noise to the concentration parameters of the prior network adaptively. 
	Study \cite{AnomalyDetection10} trained a multi-class classifier over self-labeled data created by using a variety of geometric transformations to InD images. 
	They adapted the softmax value of the InD training images as anomaly scores. 
	In \cite{AnomalyDetection11}, the authors proposed a method based on the intuition that OoD data is composed of different factors than InD data that cannot be reconstructed through the same technique as the InD data.
	
	The unsupervised methods typically compare the proposed anomaly scores with a certain threshold. 
	Hence proper thresholds based on OoD validation samples can improve the performance. 
	However, the hyper-parameters tuned by one OoD dataset may not be relevant to other samples \cite{AnomalyDetection12}. 
	The space of OoD data is usually too large to be covered, which leads to the selection bias of learning.
	In \cite{GODIN}, the authors followed the setting of \cite{OoD_detect_baseline} and proposed a dividend/divisor structure for a classifier. 
	The authors introduced a new probabilistic perspective for predicted probability confidence decomposition, especially adding a variable representing whether the data is InD or not. 
	Besides, they developed an effective strategy to tune perturbation magnitude with only InD data avoiding using OoD samples to tune the hyper-parameters. 
	
	\subsection{XAI evaluation metrics}
	Many XAI evaluation metrics emphasize quantifying characteristics that perfect XAI methods should possess, including but not limited to faithfulness, localization ability,  false-positives, sensitivity check, and stability \cite{metric}. In particular, faithfulness and localization ability are considered important properties for evaluating XAI algorithms of the saliency map kind. For example, studies \cite{simonyan2013deep, zeiler2014visualizing, zhou2016learning, petsiuk2018rise, gradcam, zhang2018top} test the algorithms’ ability to localize human-labeled objects in an image for localization. For faithfulness, the work \cite{gradcam} compares Grad-CAM with a reference saliency map, whereas \cite{petsiuk2018rise} design specific computational metrics to quantify whether XAI algorithms have faithfully assigned higher scores to more important features.  
	
	In addition to evaluate XAI methods from these computational metrics, it has been proposed that XAI methods should also be evaluated from users’ perspectives recently. More specifically, since the ultimate goal of XAI is to make human beings understand the decision logic of machine models, it can be evaluated in the aspects of mental processes related to whether users achieve a pragmatic understanding of an AI system given the XAI, such as user understanding, trust, and curiosity, etc. \cite{wang2019designing, Hoffman2018}. Most previous studies have focused on the use of user  interviews \cite{jin2021euca} or questionnaires \cite{Hoffman2018}. Some have asked participants to directly rate XAI methods, \cite{holzinger2020measuring} or compare between two XAI methods \cite{gradcam}, and examined their ratings/judgments. The choice of user study design very often depends on the characteristics of the AI system and the XAI method. Nevertheless, one common problem in previous user studies for XAI evaluation has been small sample size without clear justifications (e.g.,  \cite{kim2016examples}). This can be improved by using a larger sample size with appropriate power analysis to reduce the margin of error and increase statistical power \cite{cohen1992statistical}.

	\section{OoD-resistant perturbation-based XAI algorithm framework}\label{sec:approach}
	In this section, we first present the notations used to describe our methods, then we demonstrate the overview of our OoD-resistant perturbation-based XAI algorithm framework and detailed implementation. 
	Finally, we apply this framework to RISE \cite{petsiuk2018rise}, OCCULUSION \cite{zeiler2014visualizing} and LIME \cite{lime}. 
	The improved perturbation-based XAI algorithm framework can be applied to various types of data, such as tabular, image, and text data. 
	All of these methods follow similar steps to generate explanations. 
	This paper illustrates the framework under the image classification scenario.
	Our notation is described in Table.~\ref{table:notation}.

	\begin{table}[h]
		\caption{Notations}\label{table:notation}
		\begin{center}
			\begin{tabular}{ p{2.2cm} | p{5.8cm} }
				\hline
				\textbf{Symbol}&{\textbf{Representative}} \\
				\hline
				I & a three-dimensional RGB square image  \\
				\hline
				$S_{in}$ = $\{I\}$ & a set of input images  \\
				\hline
				$C_i$ & each possible output class label predicted by model \\
				\hline
				$N_{cls}$ & the number of all possible prediction classifications of the model  \\
				\hline
				$I \to C_i$, $i \in [1,N_{cls}]$ & the process of inference  \\
				\hline
				$p(C_i|I)$ & the prediction probability of each class $C_i$  \\
				\hline
				$M$ & each random generalized mask, a grey image   \\ 
				\hline
				$P$ & a perturbed image composed of the original image and mask \\
				\hline
				$S_m = \{M_1, M_2,$ \\ $..., M_j, ..., M_{N_{ptb}}\}$ & a set of masks \\
				\hline
				$S_p = \{P_1, P_2, ...,$ \\ $P_j, ..., P_{N_{ptb}}\}$ & corresponding set of perturbed samples  \\
				\hline
				$I \otimes S_{m} \to S_{p}$ & the process of generating a set of masks $S_m$ for one image and synthesis corresponding set of perturbation samples $S_p$   \\
				\hline
				$S$ & the final explanation output represented by saliency map  \\
				\hline
				$d_{InD}$ & domain of in distribution \\
				\hline
			\end{tabular}
			\label{tab1}
		\end{center}
	\end{table}

	\subsection{Framework}
\begin{figure*}[ht]
		\centering
		\includegraphics[width=0.9\linewidth]{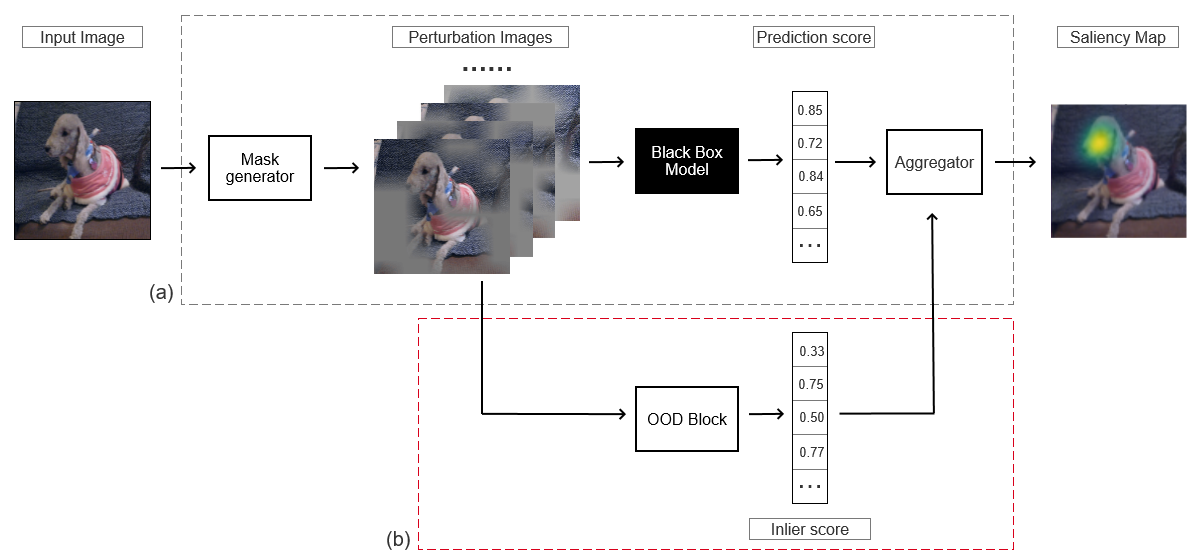}
		\caption{The general pipeline of our perturbation-based method. (a) The original pipeline. (b) The additional module we employ to address the OoD data problem.
		}
		\label{fig:pipeline}
	\end{figure*}

	Traditional perturbation-based XAI approaches usually generate a set of masks $S_m$ for one image and synthesize a corresponding set of perturbation samples $S_p$. 
	Then these perturbed samples are subsequently fed into the model for classification. 
	Different algorithms adopt different methods to calculate the final explanation saliency map. 
	The process is shown in Fig.~\ref{fig:pipeline} (a). 
	To deal with the artifact caused by the aforementioned OoD data problem, our framework integrates an OoD Block inside the whole pipeline of perturbation-based XAI methods, as is shown in Fig.~\ref{fig:pipeline} (b). 
	The OoD block is used to calculate the `Inlier Score' $s_j$ for each perturbed image $P_j$, an approximation for the probability of $P_j$ being consistent with the original dataset  distribution. 
	A perturbed image $P_j$ with a higher inlier score $s_j$ means the sample is more likely generated from the original distribution and vice versa. 
	Methods without OoD correction implicitly assume that all perturbed images belong to the original dataset distribution. 
	However, in the presence of the OoD issue, joint probability $p \left( C_i, InD \left| P_j\right. \right)$ of $P_j$ belonging to class $C_i$ and being InD simultaneously is more informative than the marginal probability $\left( C_i \left| InD, P_j\right. \right)$ per se. 
	Joint probability $p \left( C_i, InD \left| P_j\right. \right)$ can be factorised into the product of the conditional probability $p \left( InD \left| P_j\right. \right)$ and probability of the condition $p \left( C_i \left| InD, P_j\right. \right)$ with the explicit probability of InD in consideration:
	\begin{equation}
		p \left( C_i, InD \left| P_j\right. \right) =   p \left( InD \left| P_j\right. \right) \cdot p \left( C_i \left| InD, P_j\right. \right)
		\label{eq:conditional_probability}
	\end{equation}
	Given that $s_j$ is a proxy for $p\left( InD \left| P_j\right. \right)$ and model output $p(C_i|P_j)$ coincides with $p \left( C_i \left| InD, P_j\right. \right)$, the adjusted score $p(C_i|P_j) \times s_j$ gives a more accurate estimate for the probability of interest. OoD samples with a high prediction probability should be punished and reduced to adjust the over-confidence probability in order to improve the explanation performance.
	
	The saliency map $S$ is generated by aggregating scores $p(C_i|P_j) s\times s_j$ with the perturbation masks $M_j$ instead of just a combination of $p(C_i|P_j)$ and $M_j$.
	The algorithm can be summarised as follows:
	\begin{enumerate}
		\item Input image $I$ is fed to mask generator and multiplied element-wise with masks $S_m$;
		\item The resulting perturbed images $S_p$ are subsequently fed to the model for classification
		\item The model produces $p(C_i|P)$ for every perturbation image $P$.
		\item OoD block works out an inlier score $s_j$ for every perturbed image $P$.
		\item A saliency map for the original image is created based on masks($S_m$), probability scores and inlier scores. 
		\begin{equation}
			S = \mathop{Aggre}\limits_{j}(M_j, p(C_i|P_j), s_j)
			\label{eq:aggregate}
		\end{equation}
	\end{enumerate}
	
	\subsection{Inlier score calculation}
	After a comprehensive investigation in Section 2, we finally refer to G-ODIN \cite{GODIN} to design our OoD Block. 
	G-ODIN follows ODIN \cite{ODIN} and mainly comprises two strategies: input pre-processing and decomposed confidence. 
	Input pre-processing makes InD data and OoD data more distinguishable by adding some noises \cite{preprocessing}. 
	Decomposed confidence separates the distance between samples and every class center from a learnable temperature scaling parameter. 
	The straightforward target of this procedure is to learn a classifier to predict the joint probability $p \left( C_i,  InD \left| I\right. \right)$ by considering both supervisions on class $C_i$ and domain d. 
	Thus, G-ODIN uses the structure of Equation \ref{eq:conditional_probability} as prior knowledge to design classifier logits $f_i(I)$ for each class $C_i$: 
	\begin{equation}
		f_i(I) = \frac{h_i(I)}{g(I)}
		\label{eq:decompose}
	\end{equation}
	We use $h_i(I)$ and $g(I)$ to build a joint probability classifier. 
	Both $h_i(I)$ and a learnable temperature scaling module $g(I)$ are trained based on the same training set.
	A backbone classification model, such as ResNet or DenseNet, needs to be trained on the InD training dataset for the purpose of feature extraction. 
	The penultimate layer yields the features $f^p(I)$ and last layer weights $\omega_i$ are used to calculate a similarity score $h_i(I)$ between the image and each class $C_i$:
	\begin{equation}
		h_i(I) = \frac{\omega_i^T f^p(I)}{\left \|\omega_i \right\| \left\| f^p(I) \right\|}.
		\label{eq:cosine}
	\end{equation}
	Here, we adopt a cosine similarity metric as it is the most efficient in the high-dimensional case.\cite{highdimensional_curse}
	Our approach uses the maximum value of similarity $max_ih_i(I)$ between the input sample with each class center to calculate the inlier score.
	The $g(I)$ produces different temperatures for different inputs. 
	It helps to flatten the overall classification probability in the early stage of the model training process and increases cross-entropy. 
	As a consequence, it helps avoid the situation where the classifier falls into a local optimum. 
	
	The final formula for inlier score calculation is given by
	\begin{equation}
		\begin{split}
			&s_I = \left\{
			\begin{aligned}
				& 1, \quad\quad if \quad\quad max_ih_i(I) \geq max_ih_i(I)_{origin}, \\
				& 0, \quad\quad if \quad\quad max_ih_i(I) \leq max_ih_i(I)_{grey}, \\
				& \dfrac{max_ih_i(I) - max_ih_i(I)_{grey}}{max_ih_i(I)_{origin} - max_ih_i(I)_{grey}}, otherwise  
			\end{aligned}
			\right. \\ 
			& s.t. \quad\quad max_ih_i(I) \in [-1, 1]
		\end{split}
		\label{eq:normalization}    
	\end{equation}
	where $max_ih_i(I)_{grey}$ and $max_ih_i(I)_{origin}$ denote the $max_ih_i(I)$ value of the image with all pixels occluded and the image with no pixels occluded respectively. 
	We assume that the original image is a canonical example of an InD image while the purely grey image is the most representative OoD example. 
	The other perturbed images lie somewhere in between and are normalized to this range according to their $max_ih_i(I)$ value. 
	This step harness the similarity between the perturbed sample and each category center and yields a proxy for $p\left( InD \left| P_j\right. \right)$, which plays an important role in the OoD correction of the saliency map, as Equation \ref{eq:aggregate} shows.
	
\begin{figure}[h]
		\centering
		\includegraphics[scale=0.5]{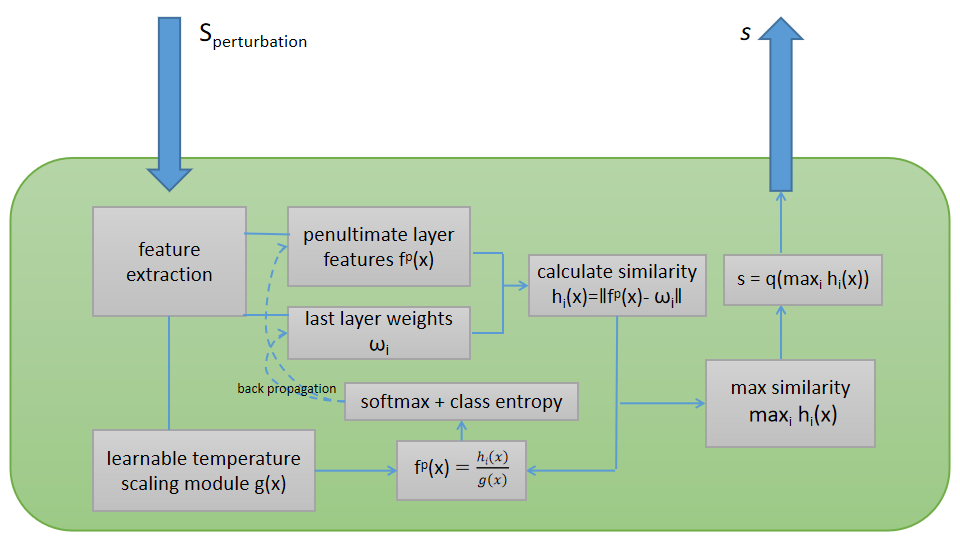}
		\caption{OoD Block structure}
		\label{fig:OOD_block}
	\end{figure}

	The structure of the OoD block is summarized in Fig.~\ref{fig:OOD_block}. Network parameters are updated by continuously optimizing the distance between the penultimate features and the category weights. 
	\vspace{-3ex}
	
	The OoD correction procedure itself is reasonably fast. Although training the backbone classification model is indeed costly, it is done only once and the original classifier does not need to be retrained for each perturbed image. 
	The only operation performed for each masked image is the similarity score calculation and normalization which does not overburden the computations. 
	The total computing time is proportional to the number of perturbed images. 
	In the experiment section, we demonstrate our empirical results as a reference.
	
	\subsection{Application to RISE}
	
	RISE \cite{petsiuk2018rise} creates a saliency map for AI models, indicating which parts of input data are the most relevant to the prediction. The method follows a simple yet powerful procedure:
	\begin{enumerate}
		\item Each input image is multiplied with a randomly generated mask that takes up part of the original image. To achieve better performance, the perturbation mask is blocky and continuous by upsampling the downscaled random mask rather than disturbing pixels directly.
		\item These synthesized images are subsequently fed to the model for classification.
		\item The model estimates probabilities $p(C_i|P_j)$ for each perturbed image $P_j$ to belong to each class $C_i$.
		\item A saliency map for the original image is created as a linear combination of the masks. The coefficients of this linear combination are probability scores produced by the model. Suppose that every perturbation sample has its predicted class $C_i$ and corresponding predicted probability. Here is how the saliency map is calculated for explaining the $C_i$:
		\begin{equation}
			S = \sum_{j=1}^{N_{ptb}} p(C_i|P_j) \times M_j 
			\label{eq:rise_sum}
		\end{equation}
	\end{enumerate}
	
\begin{figure}[h]
		\centering
		\includegraphics[scale=0.4]{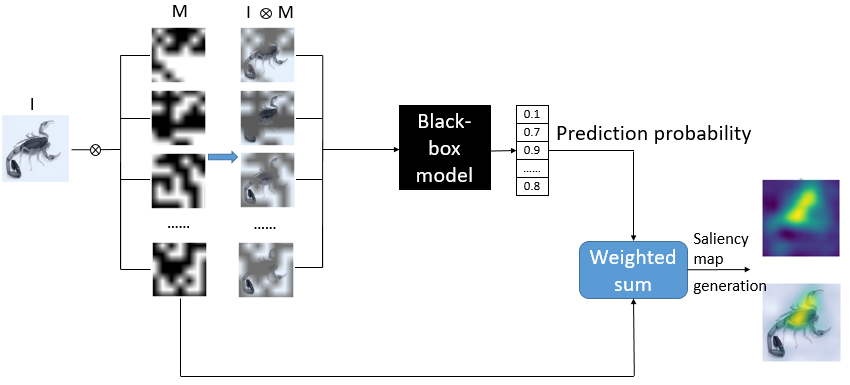}
		\caption{Algorithm flow of RISE}
		\label{fig:rise}
	\end{figure}

	The entire algorithm \cite{petsiuk2018rise} is illustrated in Fig.~\ref{fig:rise}.
	
\begin{figure}[h]
		\centering
		\includegraphics[scale=0.4]{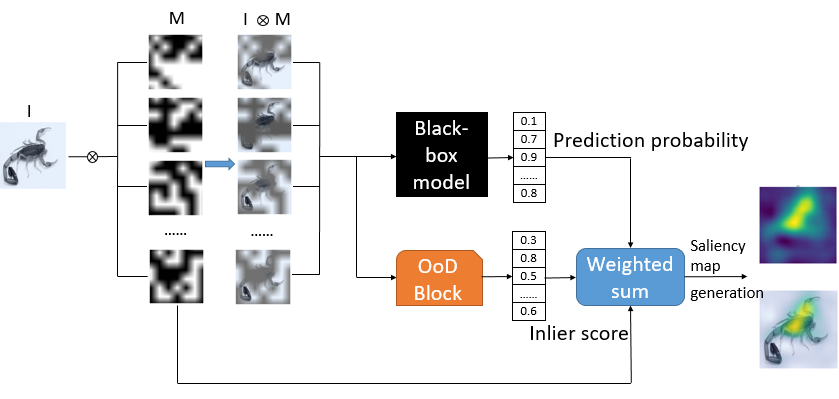}
		\caption{Algorithm flow of $RISE^+$}
		\label{fig:rise+}
	\end{figure}

	The improved version of this method $RISE^+$ contains an additional OoD Block where the inlier scores $s_j$ are calculated, as shown in Fig.~\ref{fig:rise+}.
	
	At the final step, probability scores and inlier scores are aggregated with masks, generating the resulting saliency map: 
	\begin{equation}
		S^+ = \sum_{j=1}^{N_{ptb}} p(C_i|P_j) \times s_j \times M_j.
		\label{eq:rise+_sum}
	\end{equation}
	
	\subsection{Application to OCCLUSION}
	OCCLUSION \cite{zeiler2014visualizing} measures the importance of a fixed size image region by occluding the region to examine the differences in the model’s prediction. The corresponding saliency map is generated as follows:
	\begin{enumerate}
		\item Systematically shifts a patch to occlude image $I$ by fixed stride to generate perturbed images $S_p$.
		\item The perturbed images and the original image are fed to the model for classification. For each pair of perturbation image $P_j$ and original image $I$, the model produces probability scores $p(C_i|I)$ and $p(C_i|P_j)$ for $C_i$.
		\item The difference between $p(C_i|I)$ and $p(C_i|P_j)$ is regarded as the importance of the occluded region in the perturbed image $P_j$: a substantial drop in the prediction score indicates high importance of the occluded region.
		\item The saliency map indicating the importance region on an image is generated by aggregating the differences across all pairs of images, as shown in Equation \ref{eq:occlusion+}.
		\begin{equation}
			S = \sum_{j=1}^{N_{ptb}} (p(C_i|I) - p(C_i|P_j)) \times M_j 
			\label{eq:occlusion+}
		\end{equation}
	\end{enumerate}
	Similar to $RISE^+$, the OoD Block where inlier scores $s_j$ are calculated is added to the main body of the OCCLUSION algorithm. 
	We call this improved algorithm $OCCLUSION^+$. 
	To avoid the OoD-related artifacts, we use the inlier scores to adjust the final saliency score as follows: 
	
	\begin{equation}
		S^+ = \sum_{j=1}^{N_{ptb}} ((p(C_i|I) - p(C_i|P_j) \times s_j) \times M_j 
		\label{eq:occlusion+_sum}
	\end{equation}
	
	\subsection{Application to LIME}
	In LIME \cite{lime}, the importance of each feature is obtained through the following procedure. 
	The original data is perturbed around the neighborhood, generating a set of perturbation samples $S_p$. 
	All those samples are predicted by the black-box model with $p(C_i|P_j)$. Then a simple, interpretable model, such as linear regression or a decision tree, is trained on this newly-created dataset of perturbed instances, revealing the importance of each feature as the explanation. 
	In $LIME^+$, OoD block yields the inlier score for each perturbation sample which is added to the LIME's algorithm process as well. 
	The inlier score is multiplied by $p(C_i|P_j)$ in order to adjust the predicted probabilities.
	
	\section{XAI metric analysis and improvement}\label{sec:metric}
	In this section, we re-examine existing XAI metrics both from computational and cognitive perspectives. 
	We propose `$Deletion^+$' and `$WIoSR$'  indicators and use them to quantify the advantage of our method in the experiment section. 
	We also design human study experiments based on cognition metrics as well.  
	
	\subsection{Faithfulness}
\begin{figure}[h]
		\centering
		\includegraphics[scale=0.5]{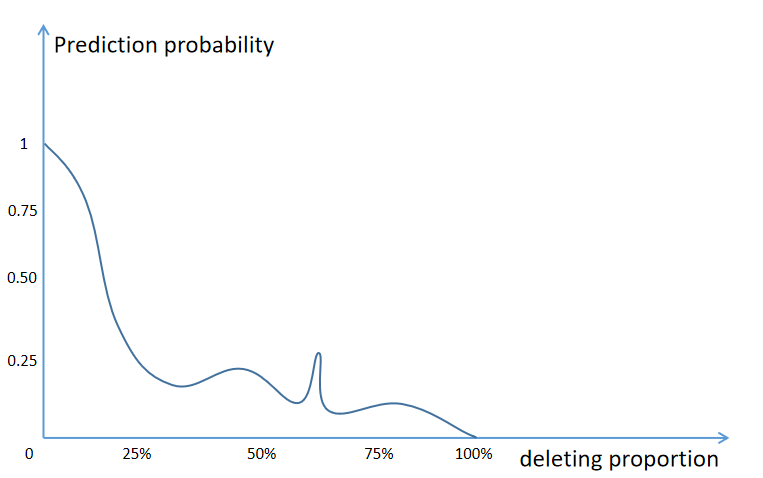}
		\caption{Deletion calculation}
		\label{fig:faithfulness}
	\end{figure}

	Faithfulness is a metric proposed in \cite{NEURIPS2018_faithfulness} indicting how much the features explained by XAI algorithms are aligned with the features used by the model for prediction. 
	It relies on the assumption that a feature is essential for classification if its removal will lead to a significant drop in the prediction probability. 
	One of the most common calculation methods is `Deletion', shown in Fig.~\ref{fig:faithfulness}. 
	\vspace{-3ex}
	\vspace{-2ex}
	Y-axis corresponds to the predicted probability values while X-axis is the percentage of content deleted. 
	Typically, the deletion order is based on the importance of each part given by XAI algorithms. 
	A better XAI algorithm points out key parts where the probabilities drop sharply. 
	This area value under the curve is faithfulness and the better the XAI method performs, the smaller this value is. 
	The below equation is used for faithfulness calculation:
	\vspace{-2ex}
	\begin{equation}
		Deletion = \int_{0}^{1} p(Ci | P_{\theta}) d\theta
		\label{eq:faithfulness}
	\end{equation}
	where $\theta$ represents the deletion proportion and $P_{\theta}$ is the image where fraction $\theta$ of the most important pixels is removed.  
	
	However, this method fails to take the OoD issue into account. 
	As $\theta$ reaches a certain level, image $P_{\theta}$ becomes OoD and probability $p(Ci | P_{\theta})$ predicted by the model is no longer reliable. 
	Similar to the previous section, this bias can be reduced by using the inlier score produced by the OoD block. 
	The improved metric is named `$Deletion^+$' and computed as follows:
	\vspace{-3ex}
	\begin{equation}
		Deletion^+ = \int_{0}^{1} p(Ci | P_{\theta}) \times s_{P_{\theta}} d\theta
		\label{eq:faithfulness+}
	\end{equation}
	\vspace{-5ex}
	\subsection{Localization}
\begin{figure}[h]
		\centering
		\includegraphics[scale=0.45]{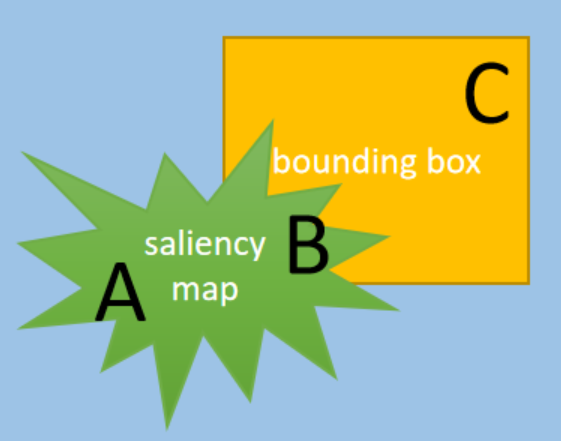}
		\caption{Illustration of a saliency map to be evaluated with localization metric}
		\label{fig:localization}
	\end{figure}

	Localization metric is widely used to evaluate XAI algorithms by checking the correspondence between the saliency map and the ground-truth bounding box \cite{metric}, as illustrated in Fig.~\ref{fig:localization}. 
	\vspace{-2ex}
	\vspace{-2ex}
	Proposed by \cite{zhou2016learning}, the localization metric assumes that black-box models make predictions based on the features from the object itself. The metric evaluates whether the explanation method correctly localizes the object to be recognized, by comparing the highlighted region in the saliency map and the bounding box of the object. XAI algorithms of the saliency maps kind give every pixel a saliency score, indicating the significance of the model prediction. Pointing Game \cite{zhang2018top}, as a naive but widely used indicator, only checks the highest score pixel whether inside the bounding box. For a particular sample, the value is 1 if hits and 0 if misses. For a group of samples, Pointing Game is computed through Equation.~\ref{eq:pointing_game}.
	\begin{equation}
		PG = \frac{\#Hits}{\#Hits + \#Misses}
		\label{eq:pointing_game}
	\end{equation}
	
	Unfortunately, it cannot distinguish different algorithms' performance in most realistic situations, even they are huge differences for human understanding \cite{metric}.
	
	The other way to define the localization metric is through calculating the ratio of intersection between the salient area and the ground truth mask over the salient region. 
	This method is called IoSR \cite{metric}. 
	IoSR is more precise than Pointing Game to some extent, but it totally ignores the brightness of the pixels. 
	For a finer measurement, we use pixels' saliency scores as their weights while calculating the localization score. We call this indicator weighted IoSR (WIoSR):
	\begin{equation}
		WIoSR = \int_{0}^{1} \frac{\sum w_b}{\sum {w_a} + \sum{w_b}} d\theta \quad s.t. \quad a \in A, b\in B 
		\label{eq:localization}
	\end{equation}
	where $w_t$ is the saliency score of pixel $t$. $\theta$ is the boundary of the saliency score that which pixels are selected to be the saliency map. The saliency score less than $\theta$ is not regarded as part of saliency map. Instead of setting a $\theta$ value by ourselves, we compute the calculus for a more comprehensive consideration. 
	As Fig.~\ref{fig:localization} shows, $A(\theta)$ is the set of all pixels with saliency score above $\theta$ that do not belong to the ground-truth bounding box while $B(\theta)$ is the set of pixels with saliency score above $\theta$ that lie inside the bounding box (overlapping parts). 
	The blue part represents an image with the yellow part of the ground-truth bounding box($B(\theta) \cup C$). XAI algorithms figure out saliency maps($A(\theta) \cup B(\theta)$) as the green part. 
	
	This index is quite intuitive: if the saliency map matches the ground truth area of importance, set $A(\theta)$ is always empty and all pixels in $B(\theta)$ have the highest saliency score. On the contrary, the perfect mismatch means that $B(\theta)$ is empty. Thus, large values of the WIoSR index are indicative of good performance while small values correspond to poor performance. 
	
	\subsection{Cognitive metric}
\begin{figure}[h]
		\centering
		\includegraphics[scale=0.56]{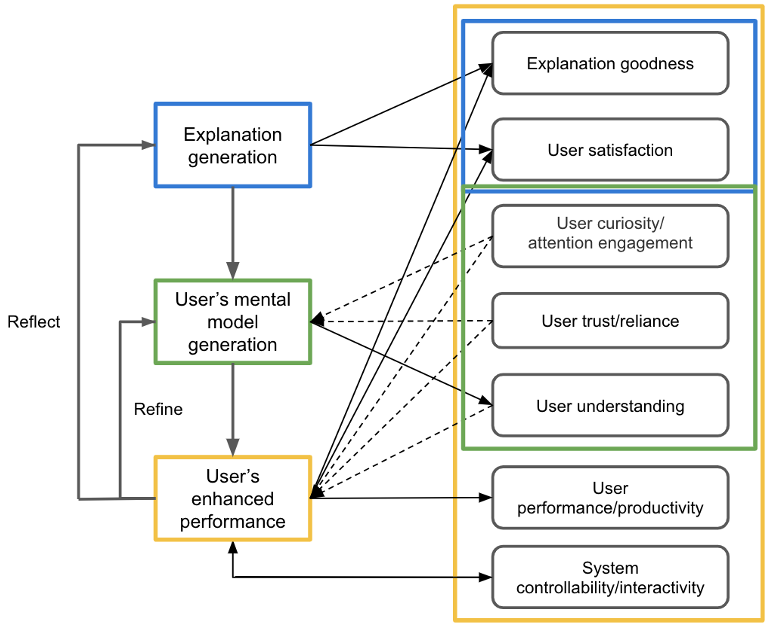}
		\caption{Hypothesized evaluative model for XAI, including three functional stages of an explanation process and seven associated cognitive metrics. The solid lines represent metrics that can be assessed at the stage, and the dashed lines represent metrics that moderate the stage.}
		\label{fig:cog_metric}
	\end{figure}

	The ultimate goal of XAI is to make human beings understand the decision logic of machine models. Consequently, the evaluation of XAI algorithms ought to not only be computational indexes but also whether users achieve a pragmatic understanding of an AI system given the XAI \cite{MILLER20191}. To evaluate the XAI systems from a cognitive science perspective, we considered three functional stages of an explanation process \cite{Hoffman2018}: XAI explanation generation, user’s mental model generation, and user’s enhanced performance. Hsiao et al. \cite{Hsiao2021} identified 7 cognitive metrics associated with these 3 stages. Here we focused on 5 of them: Explanation goodness, User satisfaction, User curiosity/attention engagement, User trust/reliance, and User understanding (Fig.~\ref{fig:cog_metric}). We conducted a user study to compare saliency maps generated from baseline methods and our improved methods along with these 5 cognitive metrics.
	
	\section{EXPERIMENTS}\label{sec:experiments}
	In this section, we report our experimental study on comparing three improved algorithms $RISE^+$, $OCCLUSION^+$, $LIME^+$ derived from our framework with baseline solutions. The evaluation is conducted with both computational metrics and cognitive metrics.
	
	\subsection{Experiment setup}
	
	\begin{table*}[ht]
		\caption{Questionnaire}
		\centering
		\label{table:questionaire}
		\begin{tabular}{|l|l|l|}
			\hline
			No. & Question                                                                                                              & Cognitive metric evaluated                                             \\ \hline
			1   & Which XAI highlights more accurate details?                                                                           & Explanation goodness                  \\ \hline
			2   & Which XAI highlights a more appropriate amount of details?                                                            & Explanation goodness   \\ \hline
			3   & Which XAI highlights fewer irrelevant details?                                                                        & Explanation goodness                  \\ \hline
			4   & Which XAI has a more reasonable size of highlighted region?                                                           & Explanation goodness                  \\ \hline
			5   & Which XAI seems more complete?                                                                                        & Explanation goodness   \\ \hline
			6   & Which XAI are you more satisfied with?                                                                                & User satisfaction                     \\ \hline
			7   & Which XAI is more reliable and you can count on more?                                                                 & User trust/reliance                            \\ \hline
			8   & \begin{tabular}[c]{@{}l@{}}Which XAI provides explanations that make you want to know more about\\ how the AI system works?\end{tabular} & User curiosity/attention engagement \\ \hline
			9   & Which XAI gives you more confidence in the AI system that it works well?                                              & User trust/reliance                                 \\ \hline
			10  & Which XAI facilitates your understanding of AI system more?                                                           & User understanding                         \\ \hline
		\end{tabular}
	\end{table*}
	
	\subsubsection{Common setting}
	We use pre-trained deep neural network models ResNet50 \cite{he2016deep} and VGG16 \cite{vgg16} as classifiers that need to explain. 
	For the OoD block, we use a ResNet-50 network as the backbone for feature extraction. 
	The hyper-parameters are set as follows: we set the batch size as 512 and the number of epochs as 200. 
	We choose SGD as the optimizer with a learning rate initialized with 0.1 and decaying by 10 every 30 epochs. 
	The OoD block is trained on four GPUs. It takes approximately 68 minutes and 144 hours to train the OoD block for VOC and ImageNet datasets, respectively. 
	These settings about the OoD block are the same with the reference anomaly detection algorithm G-ODIN \cite{GODIN}.
	
	\subsubsection{Dataset} This paper considers two publically available benchmark datasets (i.e. ImageNet \cite{russakovsky2015imagenet} and Pascal VOC \cite{voc}) for image classification. 
	ImageNet provides ILSVRC (ImageNet Large Scale Visual Recognition Challenge)\footnote{www.image-net.org/challenges/LSVRC/}, which is an annual competition that contains subsets from the ImageNet dataset. 
	We use the train/validation data from ILSVRC2012 and VOC2007. 
	Moreover, both two datasets already have bounding box annotations that can be used as ground truth for object localization in our experiments.
	The statistic analysis of datasets is shown in Table.~\ref{table:statistics}. 
	
	\begin{table}[h]
		\centering
		\caption{Statistics of experimental dataset.}       \label{table:statistics}
		\begin{tabular}{|c|c|c|c|}
			\hline
			Dataset                   &         & Training set & Validation set \\ \hline
			\multirow{2}{*}{ImageNet} & Images  & 1.2 million  & 50,000         \\ \cline{2-4} 
			& Classes & 1,000        & 1,000          \\ \hline
			\multirow{2}{*}{VOC}      & Images  & 4510         & 502            \\ \cline{2-4} 
			& Classes & 20           & 20             \\ \hline
		\end{tabular}
	\end{table}
	
	For the computational metrics experiments, we select 500 images as one group and repeat the experiment for multiple times with different group of data.
	For cognitive metrics experiments, considering the limited concentration time of the subjects, the randomly selected stimuli included 30 sets of images, with each set containing one original stimulus for AI classification, and two XAI algorithms' output in the form of saliency maps. Each algorithm contains 10 sets of images. The two XAI algorithms' explanations were generated from baseline methods and our improved methods, respectively.
	
	\subsubsection{Computational Metric}
	We evaluate the performance of the proposed algorithms by using both computational and cognitive metrics. In computational comparison, two faithfulness metrics (Deletion and $Deletion^+$) and two localization metrics (Pointing Game and WIoSR) are adopted, which are elaborated in Section \uppercase\expandafter{\romannumeral4}. The small value of faithfulness corresponds to high XAI performance, while high values are expected with the localization. We conducted the same experiment several times and reported statistically significant results and bold the results that are advantageous in the following tables. 
	
	\subsection{Cognitive metric user study design}
\begin{figure}[h]
		\centering
		\includegraphics[scale=0.13]{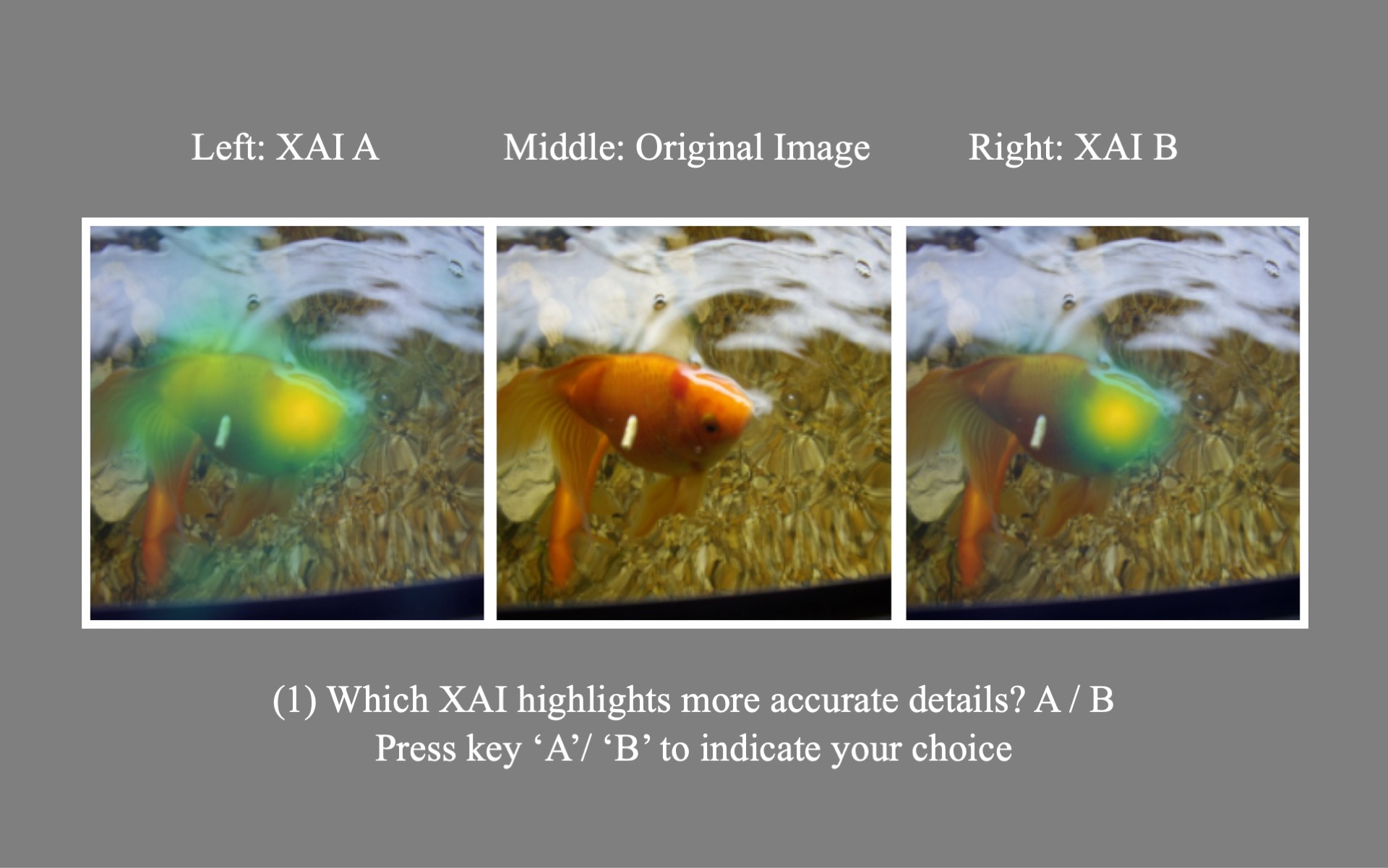}
		\caption{An example of the frame with an original image and 2 XAI methods in the saliency map format}
		\label{fig:questionaire}
	\end{figure}

	We conducted a user study to evaluate the methods using cognitive metrics. Participants were asked to judge between two saliency maps which one was a better XAI along 5 cognitive metrics:  Explanation goodness, User satisfaction, User curiosity/attention engagement, User trust/reliance, and User understanding. More specifically, in each trial two saliency maps, one generated from the baseline methods and the other from our improved methods, were presented on the left and right of the original image respectively (Fig.~\ref{fig:questionaire}) with their positions randomly determined. Three studies were conducted: RISE vs. $RISE^+$, OCCLUSION vs. $OCCLUSION^+$, and LIME VS. $LIME^+$. In each study, 10 images were randomly selected from those showing visible differences between the two XAI methods. For each image, participants answered 10 questions to choose between the two XAI methods according to the 5 cognitive metrics. For each cognitive metric, we calculated the percentage of trials each XAI method was selected. We then performed paired t-tests to examine whether the two XAI methods differed significantly in the percentage of trials selected as a better XAI. We recruited 39 adult participants for the user study. According to power analysis based on paired t-test, the minimum sample size assuming a medium effect size d = 0.50, $\alpha$ = 0.05, power = 0.8, is 34. Participants reported to have normal or corrected-to-normal vision with no cognitive disabilities or psychological problems. 
	
	To evaluate and compare two XAI methods using the current experiment setup, for each image participants answer 10 questions, each corresponding to a cognitive metric as shown in Table.~\ref{table:questionaire}. These questions were adapted from Explanation Goodness Checklist, Explanation Satisfaction Scale, Curiosity Checklist \cite{Hoffman2018}, the Recommended Scale  \cite{Hoffman2018, cahour2009d, jian2000foundations, schaefer2013perception, madsen2000measuring} and the System Causability Scale \cite{holzinger2020measuring}.
	Detail questions are listed in Table.~\ref{table:questionaire}
	
	\subsection{RISE vs $RISE^+$}
	$RISE^+$ uses the same default hyper-parameters with RISE \cite{petsiuk2018rise}. 
	Randomly selected images from ImageNet and VOC datasets are used to compare the effectiveness of RISE and $RISE^+$ from the perspective of faithfulness and localization, as shown in Table.~\ref{table:compare_rise_faithfulness} and Table.~\ref{table:compare_rise_localization}. 
	These data illustrates that $RISE^+$ outperforms RISE in all four metrics on the ImageNet dataset. In the improved metrics (Deletion+ and WIoSR), the advantages of our algorithms are more significant. 
	
	\begin{table*}[]
		\centering
		\caption{Faithfulness of RISE vs $RISE^+$ on general case}
		\label{table:compare_rise_faithfulness}
		\begin{tabular}{|c|c|l|l|l|l|l|l|}
			\hline
			\multirow{2}{*}{Dataset}  & \multirow{2}{*}{Classifier} & \multicolumn{3}{c|}{Deletion}                                                               & \multicolumn{3}{c|}{Deletion+}                                                              \\ \cline{3-8} 
			&                             & \multicolumn{1}{c|}{RISE} & \multicolumn{1}{c|}{$RISE^+$} & \multicolumn{1}{c|}{Gain}       & \multicolumn{1}{c|}{RISE} & \multicolumn{1}{c|}{$RISE^+$} & \multicolumn{1}{c|}{Gain}       \\ \hline
			
			\multirow{2}{*}{ImageNet} & ResNet50                    & 0.116$\pm$0.003           & 0.113$\pm$0.002            & \textbf{-3.12\%}             & 0.075$\pm$0.003           & 0.070$\pm$0.001               & \textbf{-7.27\%}             \\ \cline{2-8} 
			& VGG16                       & 0.113$\pm$0.003           & 0.112$\pm$0.001            & \textbf{-1.16\%}             & 0.073$\pm$0.002           & 0.069$\pm$0.000               & \textbf{-5.35\%}             \\ \hline
			\multirow{2}{*}{VOC}      & ResNet50                    & 0.187$\pm$0.006           & 0.212$\pm$0.002            & 12.91\%                      & 0.120$\pm$0.006           & 0.083$\pm$0.001               & \textbf{-30.84\%}            \\ \cline{2-8} 
			& VGG16                       & 0.106$\pm$0.002           & 0.136$\pm$0.001            & 28.58\%                      & 0.072$\pm$0.001           & 0.059$\pm$0.001               & \textbf{-17.88\%}            \\ \hline
		\end{tabular}
	\end{table*}
	
	\begin{table*}[]
		\centering
		\caption{Localization of RISE vs $RISE^+$ on general case}
		\label{table:compare_rise_localization}
		\begin{tabular}{|c|c|l|l|l|l|l|l|}
			\hline
			\multirow{2}{*}{Dataset}  & \multirow{2}{*}{Classifier} & \multicolumn{3}{c|}{Pointing Game}                                                          & \multicolumn{3}{c|}{WIoSR}                                                                  \\ \cline{3-8} 
			&                             & \multicolumn{1}{c|}{RISE} & \multicolumn{1}{c|}{$RISE^+$} & \multicolumn{1}{c|}{Gain}          & \multicolumn{1}{c|}{RISE} & \multicolumn{1}{c|}{$RISE^+$} & \multicolumn{1}{c|}{Gain}          \\ \hline
			
			\multirow{2}{*}{ImageNet} & ResNet50                    & 0.958$\pm$0.008           & 0.970$\pm$0.001               & \textbf{1.22\%}           & 0.811$\pm$0.004           & 0.831$\pm$0.001            & \textbf{2.45\%}              \\ \cline{2-8} 
			& VGG16                       & 0.978$\pm$0.005           & 0.981$\pm$0.003               & \textbf{0.33\%}           & 0.828$\pm$0.002           & 0.844$\pm$0.001            & \textbf{1.96\%}              \\ \hline
			\multirow{2}{*}{VOC}      & ResNet50                    & 0.876$\pm$0.019           & 0.888$\pm$0.002               & \textbf{1.45\%}           & 0.657$\pm$0.012           & 0.674$\pm$0.001            & \textbf{2.48\%}              \\ \cline{2-8} 
			& VGG16                       & 0.953$\pm$0.009           & 0.940$\pm$0.002               & -1.40\%                  & 0.726$\pm$0.004            & 0.724$\pm$0.001            & -0.31\%                      \\ \hline
		\end{tabular}
	\end{table*}
	
\begin{figure}[ht]
		\centering
		\includegraphics[scale=0.22]{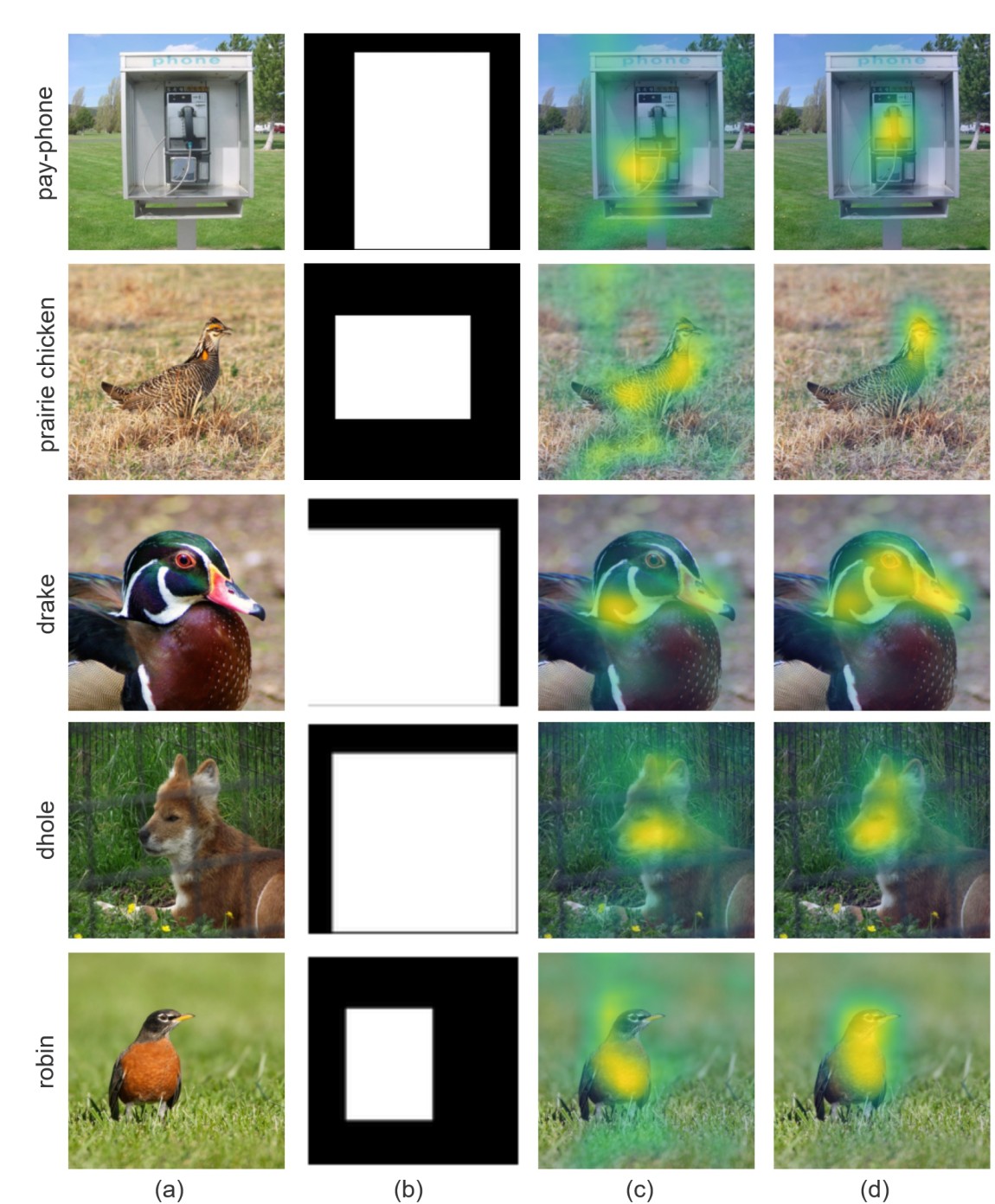}
		\caption{Evaluate RISE and $RISE^+$ on general case. (a) is the original image. (b) is the ground truth bounding box. (c) is the RISE results(composite saliency map). (d) is the $RISE^+$ results(composite saliency map).}
		\label{fig:general_rise}
	\end{figure}

	We also select parts of data to provide an intuitive understanding in Fig.~\ref{fig:general_rise}, from which we can see that the highlighted position of our improved algorithm is more discriminative (see `pay-phone' (d)), more concentrated (see `prairie chicken' (d)), and more complete (see `drake' `dhole' `robin' (d)).
	
\begin{figure*}[ht]
		\centering
		\includegraphics[width=1.0\linewidth]{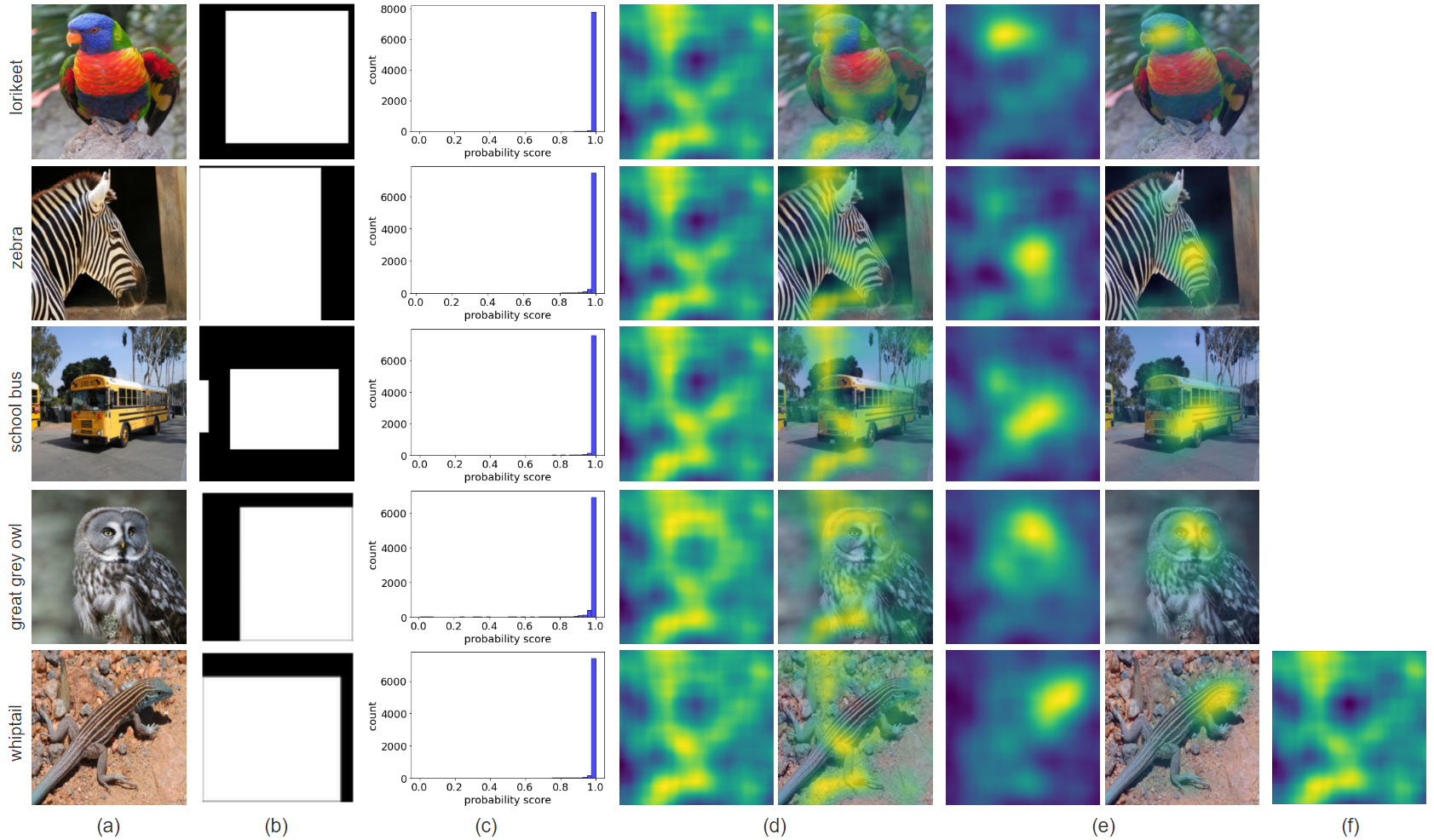}
		\caption{Evaluate RISE and $RISE^+$ on degradation case. (a) is the original image. (b) is the ground truth bounding box. (c) is model prediction distribution graph. (d) is the RISE result(separate and composite saliency map). (e) is the $RISE^+$ results(separate and composite saliency map). (f) is the average mask}
		\label{fig:degradation_rise}
	\end{figure*}

	Specially, we found there is one `degradation case' in RISE. `Degradation' refers to the condition that no matter which image is to explain, RISE gives the almost same saliency map as the explanation. The reason of degradation is RISE inherent bias from mask generation. Some parts of the image are covered more frequently than others, and hereby more likely to be highlighted in the result saliency map. We note that we would obtain an image of the `average mask' by combining all masks together with the same weights, shown in Fig.~\ref{fig:degradation_rise} (f). When degradation is severe, all results would be degraded to the "average mask" and therefore make the explanation meaningless. We compute the `Pearson Correlation Coefficient' between result saliency maps and the `average mask' and define the case whose range above 0.8 is `degradation', which is regarded as the two images' pattern are highly correlate. Generally, this case occurs when no matter which part is masked, the prediction probabilities are almost the same. (e.g. the target is almost full of the whole image). Statically, the occurrence rate of degradation is 7.4\% $\pm$ 0.05 according to our experimental observation. In the degradation case, $RISE^+$ has a much greater performance than RISE. The quantitative and qualitative results are shown in Table.~\ref{table:compare_rise_degradation_faithfulness}, Table.~\ref{table:compare_rise_degradation_localization} and Fig.~\ref{fig:degradation_rise}. In the degradation case, $RISE^+$ has a significant improvement in various indicators over RISE. For clarity, we show the separate saliency map in Fig.~\ref{fig:degradation_rise}(d) as well. Every degradation case image has the almost same saliency map (average mask), which does not make sense. By contrast, $RISE^+$ gives reasonable explanation in Fig.~\ref{fig:degradation_rise}(e).

	\begin{table*}[ht]
		\centering
		\caption{Faithfulness of RISE vs $RISE^+$ on degradation case}
		\label{table:compare_rise_degradation_faithfulness}
		\begin{tabular}{|c|c|l|l|l|l|l|l|}
			\hline
			\multirow{2}{*}{Dataset}  & \multirow{2}{*}{Classifier} & \multicolumn{3}{c|}{Deletion}                                                               & \multicolumn{3}{c|}{Deletion+}                                                              \\ \cline{3-8} 
			&                             & \multicolumn{1}{c|}{RISE} & \multicolumn{1}{c|}{$RISE^+$} & \multicolumn{1}{c|}{Gain}       & \multicolumn{1}{c|}{RISE} & \multicolumn{1}{c|}{$RISE^+$} & \multicolumn{1}{c|}{Gain}       \\ \hline
			\multirow{2}{*}{ImageNet} & ResNet50                    & 0.411$\pm$0.033           & 0.354$\pm$0.021            & \textbf{-13.79\%}            & 0.256$\pm$0.028           & 0.192$\pm$0.017            & \textbf{-25.04\%}            \\ \cline{2-8} 
			& VGG16                       & 0.398$\pm$0.039           & 0.372$\pm$0.023            & \textbf{-6.56\%}             & 0.260$\pm$0.032           & 0.219$\pm$0.017            & \textbf{-15.77\%}            \\ \hline
			\multirow{2}{*}{VOC}      & ResNet50                    & 0.564$\pm$0.060           & 0.504$\pm$0.025            & \textbf{-10.61\%}            & 0.356$\pm$0.047           & 0.147$\pm$0.004            & \textbf{-58.55\%}            \\ \cline{2-8} 
			& VGG16                       & 0.505$\pm$0.067           & 0.528$\pm$0.016            & 4.54\%                       & 0.494$\pm$0.269           & 0.151$\pm$0.013            & \textbf{-69.33\%}            \\ \hline
		\end{tabular}
	\end{table*}
	
	\begin{table*}[h]
		\centering
		\caption{Localization of RISE vs $RISE^+$ on degradation case}
		\label{table:compare_rise_degradation_localization}
		\begin{tabular}{|c|c|l|l|l|l|l|l|}
			\hline
			\multirow{2}{*}{Dataset}  & \multirow{2}{*}{Classifier} & \multicolumn{3}{c|}{Pointing Game}                                                          & \multicolumn{3}{c|}{WIoSR}                                                                  \\ \cline{3-8} 
			&                             & \multicolumn{1}{c|}{RISE} & \multicolumn{1}{c|}{$RISE^+$} & \multicolumn{1}{c|}{Gain}          & \multicolumn{1}{c|}{RISE} & \multicolumn{1}{c|}{$RISE^+$} & \multicolumn{1}{c|}{Gain}          \\ \hline
			
			\multirow{2}{*}{ImageNet} & ResNet50                    & 0.794$\pm$0.059           & 0.952$\pm$0.000            & \textbf{20.00\%}             & 0.755$\pm$0.032           & 0.883$\pm$0.007            & \textbf{16.85\%}             \\ \cline{2-8} 
			& VGG16                       & 0.897$\pm$0.074           & 1.000$\pm$0.000            & \textbf{11.50\%}             & 0.785$\pm$0.030           & 0.881$\pm$0.017            & \textbf{12.27\%}             \\ \hline
			\multirow{2}{*}{VOC}      & ResNet50                    & 0.527$\pm$0.131           & 0.926$\pm$0.019            & \textbf{75.84\%}             & 0.562$\pm$0.053           & 0.806$\pm$0.023            & \textbf{43.36\%}             \\ \cline{2-8} 
			& VGG16                       & 0.720$\pm$0.311           & 1.000$\pm$0.000            & \textbf{38.84\%}             & 0.606$\pm$0.081           & 0.868$\pm$0.021            & \textbf{43.09\%}             \\ \hline
		\end{tabular}
	\end{table*}

\begin{figure}[ht]
		\centering
		\includegraphics[scale=0.3]{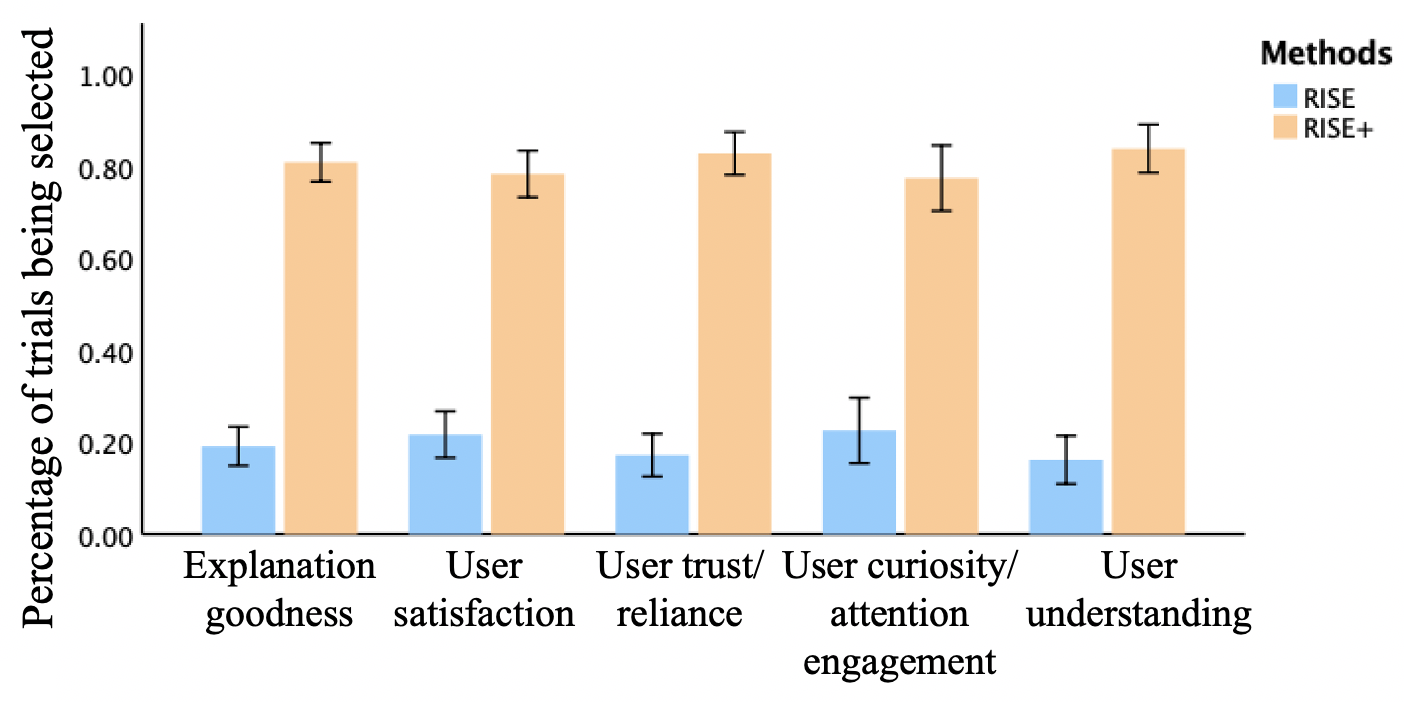}
		\caption{Percentage of trials that RISE or $RISE^+$ was selected to be a better method in terms of the 5 cognitive metrics (Error bars show one SD).}
		\label{fig:rise_cog}
	\end{figure}

	Users evaluated saliency maps generated by $RISE^+$ to be significantly better than those generated by RISE concerning explanation goodness, t(38) = 14.78, p < .001, d = 2.37, user satisfaction, t(38) = 11.33, p < .001, d = 1.81, user curiosity/ attention engagement, t(38) = 7.80, p < .001, d = 1.25, user trust/reliance, t(38) = 14.27, p < .001, d = 2.29, and user understanding, t(38) = 13.08, p < .001, d = 2.09. The comparison result is shown in Fig.~\ref{fig:rise_cog}.
	
	\subsection{OCCLUSION vs $OCCLUSION^+$}
\begin{figure}[]
		\centering
		\includegraphics[scale=0.2]{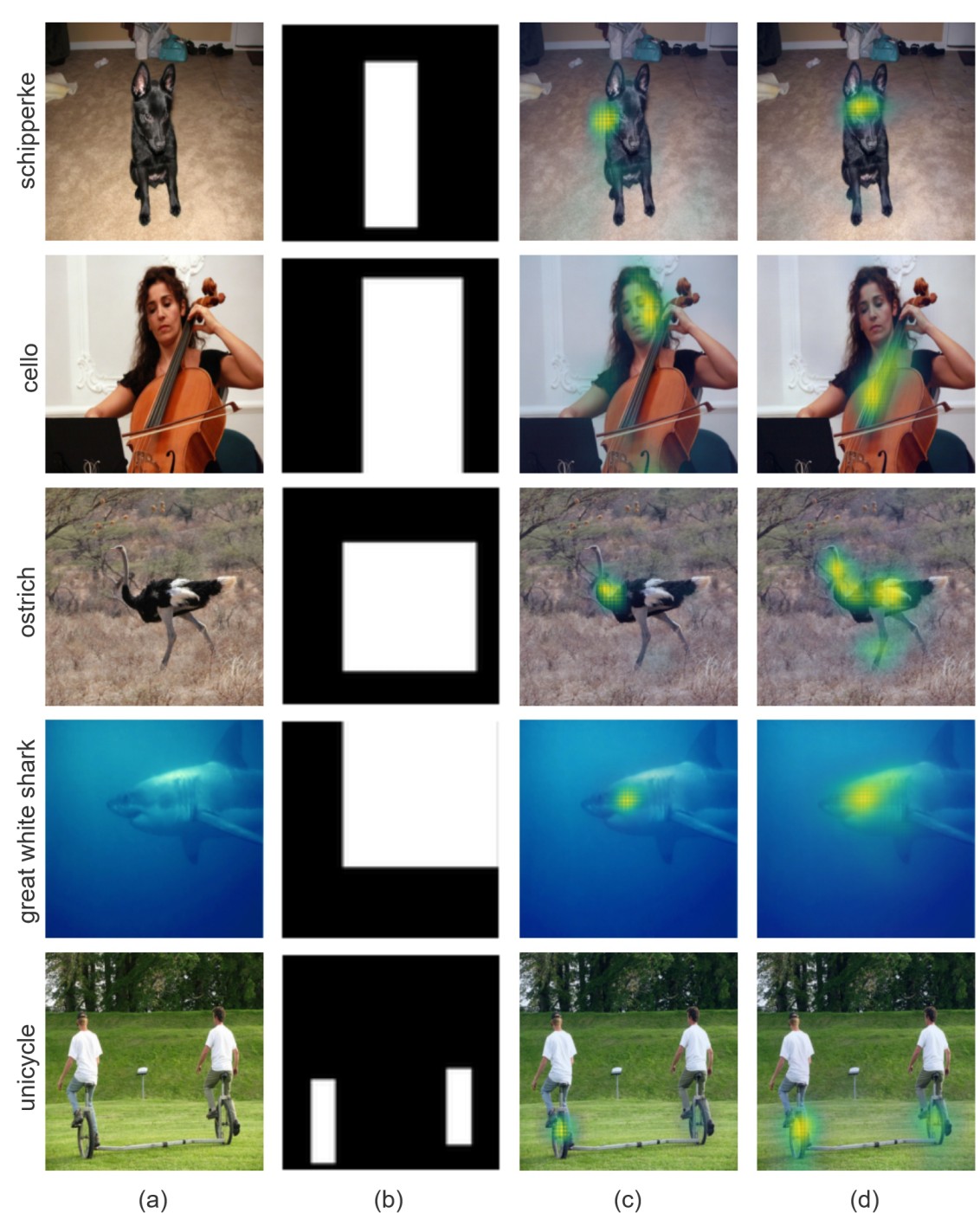}
		\caption{Evaluate OCCLUSION and $OCCLUSION^+$. (a) is the original image. (b) is the ground truth bounding box. (c) is the OCCLUSION results(composite saliency map). (d) is the $OCCLUSION^+$ results(composite saliency map).}
		\label{fig:occlusion}
	\end{figure}

	For OCCLUSION and $OCCLUSION^+$, we set the width and height of sliding patch to occlude image to 50, and the stride to be 3. Hence, 2025 masks are generated for the image size (i.e., $224\times224$). Experiments results are as Table.~\ref{table:compare_occlusion_faithfulness}, Table.~\ref{table:compare_occlusion_localization} and Fig.~\ref{fig:occlusion}. For OCCLUSION, our advantage is mainly reflected in the Deletion+ indicator. Similar to the previous comparison group, Fig.~\ref{fig:occlusion} shows that $OCCLUSION^+$ has a better positioning accuracy (see `schipperke' `cello' (d)) and highlight completeness (see `ostrich' `great white shark') than OCCLUSION. For the case of multi-target, $OCCLUSION^+$ marks all of them while OCCLUSION only identifies one of them (see `unicycle' (d)).
	
	\begin{table*}[bp]
		\centering
		\caption{Faithfulness of OCCLUSION vs $OCCLUSION^+$}
		\label{table:compare_occlusion_faithfulness}
		\begin{tabular}{|c|c|l|l|l|l|l|l|}
			\hline
			\multirow{2}{*}{Dataset}  & \multirow{2}{*}{Classifier} & \multicolumn{3}{c|}{Deletion}                                                                         & \multicolumn{3}{c|}{Deletion+}                                                                        \\ \cline{3-8} 
			&                             & \multicolumn{1}{c|}{OCCLUSION} & \multicolumn{1}{c|}{$OCCLUSION^+$} & \multicolumn{1}{c|}{Gain}          & \multicolumn{1}{c|}{OCCLUSION} & \multicolumn{1}{c|}{$OCCLUSION^+$} & \multicolumn{1}{c|}{Gain}          \\ \hline
			
			\multirow{2}{*}{ImageNet} & ResNet50                    & 0.179$\pm$0.015 & 0.152$\pm$0.009 & \textbf{-14.99\%} & 0.111$\pm$0.008 & 0.084$\pm$0.004 & \textbf{-24.01\%} \\ \cline{2-8} 
			& VGG16                       & 0.125$\pm$0.006 & 0.134$\pm$0.010 & 7.17\%            & 0.080$\pm$0.004 & 0.075$\pm$0.004 & \textbf{-5.99\%}  \\ \hline
			\multirow{2}{*}{VOC}      & ResNet50                    & 0.183$\pm$0.003 & 0.251$\pm$0.005 & 37.35\%           & 0.113$\pm$0.002 & 0.104$\pm$0.005 & \textbf{-7.95\%}  \\ \cline{2-8} 
			& VGG16                       & 0.098$\pm$0.05  & 0.159$\pm$0.012 & 62.49\%           & 0.348$\pm$0.397 & 0.346$\pm$0.398 & \textbf{-0.58\%}  \\ \hline 
		\end{tabular}
	\end{table*}
	
	\begin{table*}[bp]
		\centering
		\caption{Localization of OCCLUSION vs $OCCLUSION^+$}
		\label{table:compare_occlusion_localization}
		\begin{tabular}{|c|c|l|l|l|l|l|l|}
			\hline
			\multirow{2}{*}{Dataset}  & \multirow{2}{*}{Classifier} & \multicolumn{3}{c|}{Pointing Game}                                                                    & \multicolumn{3}{c|}{WIoSR}                                                                            \\ \cline{3-8} 
			&                             & \multicolumn{1}{c|}{OCCLUSION} & \multicolumn{1}{c|}{$OCCLUSION^+$} & \multicolumn{1}{c|}{Gain}          & \multicolumn{1}{c|}{OCCLUSION} & \multicolumn{1}{c|}{$OCCLUSION^+$} & \multicolumn{1}{c|}{Gain}          \\ \hline
			\multirow{2}{*}{ImageNet} & ResNet50                    & 0.933$\pm$0.006 & 0.950$\pm$0.008 & \textbf{1.76\%}  & 0.812$\pm$0.017 & 0.811$\pm$0.001 & -0.07\%  \\ \cline{2-8} 
			& VGG16                       & 0.935$\pm$0.008 & 0.947$\pm$0.008 & \textbf{1.25\%}  & 0.824$\pm$0.004 & 0.814$\pm$0.006 & -1.20\%  \\ \hline
			\multirow{2}{*}{VOC}      & ResNet50                    & 0.855$\pm$0.005 & 0.807$\pm$0.011 & -5.53\% & 0.702$\pm$0.008 & 0.622$\pm$0.009 & -11.45\% \\ \cline{2-8} 
			& VGG16                       & 0.881$\pm$0.044 & 0.841$\pm$0.049 & -4.49\% & 0.755$\pm$0.009 & 0.727$\pm$0.051 & -3.66\%  \\ \hline
		\end{tabular}
	\end{table*}
	
\begin{figure}[]
		\centering
		\includegraphics[scale=0.3]{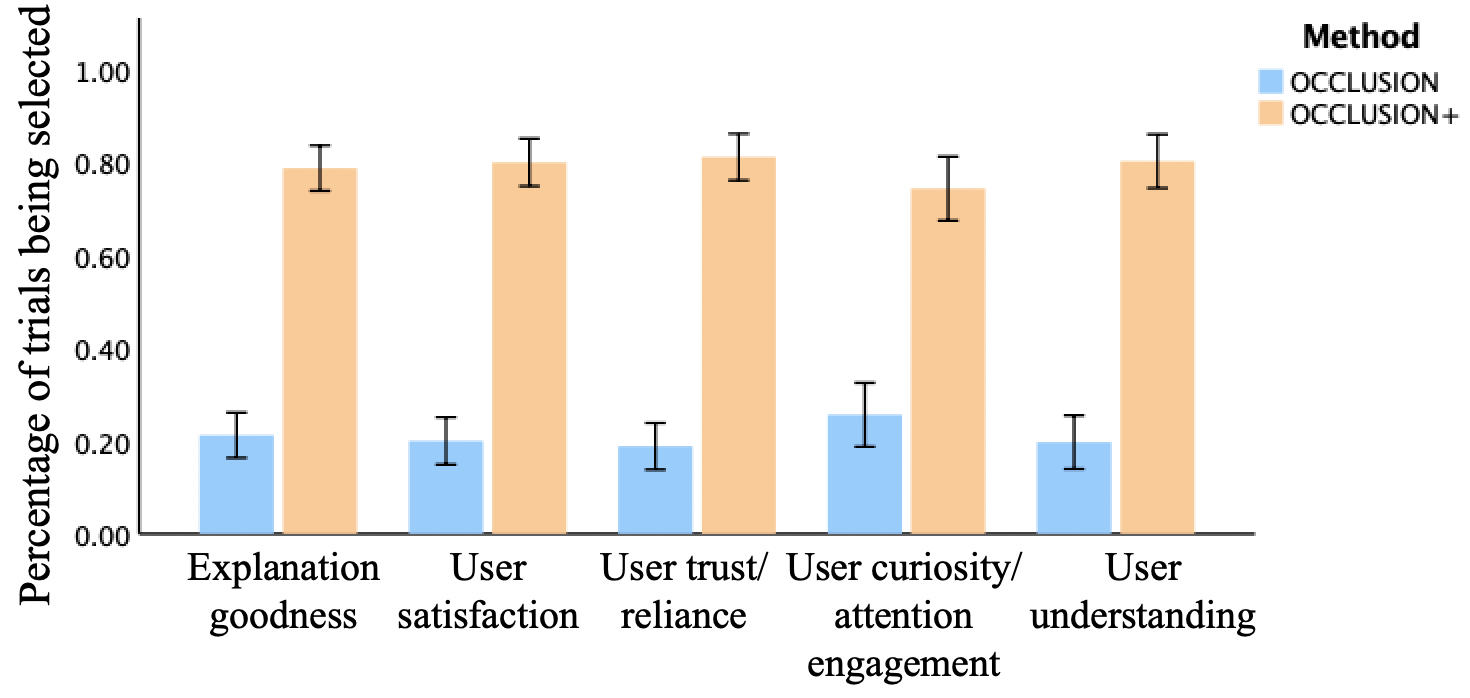}
		\caption{Percentage of trials that OCCLUSION or $OCCLUSION^+$ was selected to be a better method in terms of the 5 cognitive metrics (Error bars show one SD)}
		\label{fig:occlusion_cog}
	\end{figure}

	Similarly, users evaluated saliency maps using $OCCLUSION^+$ to be significantly better than those using OCCLUSION concerning explanation goodness, t(38) = 11.92, p < .001, d = 1.91, user satisfaction, t(38) = 11.88, p < .001, d = 1.90, user curiosity/ attention engagement, t(38) = 7.16, p < .001, d = 1.15, user trust/reliance, t(38) = 12.56, p < .001, d = 2.01, and user understanding, t(38) = 10.59, p < .001, d = 1.70. Please refer to Fig.~\ref{fig:occlusion_cog}.
	
	\subsection{LIME vs $LIME^+$}
\begin{figure}[]
		\centering
		\includegraphics[scale=0.21]{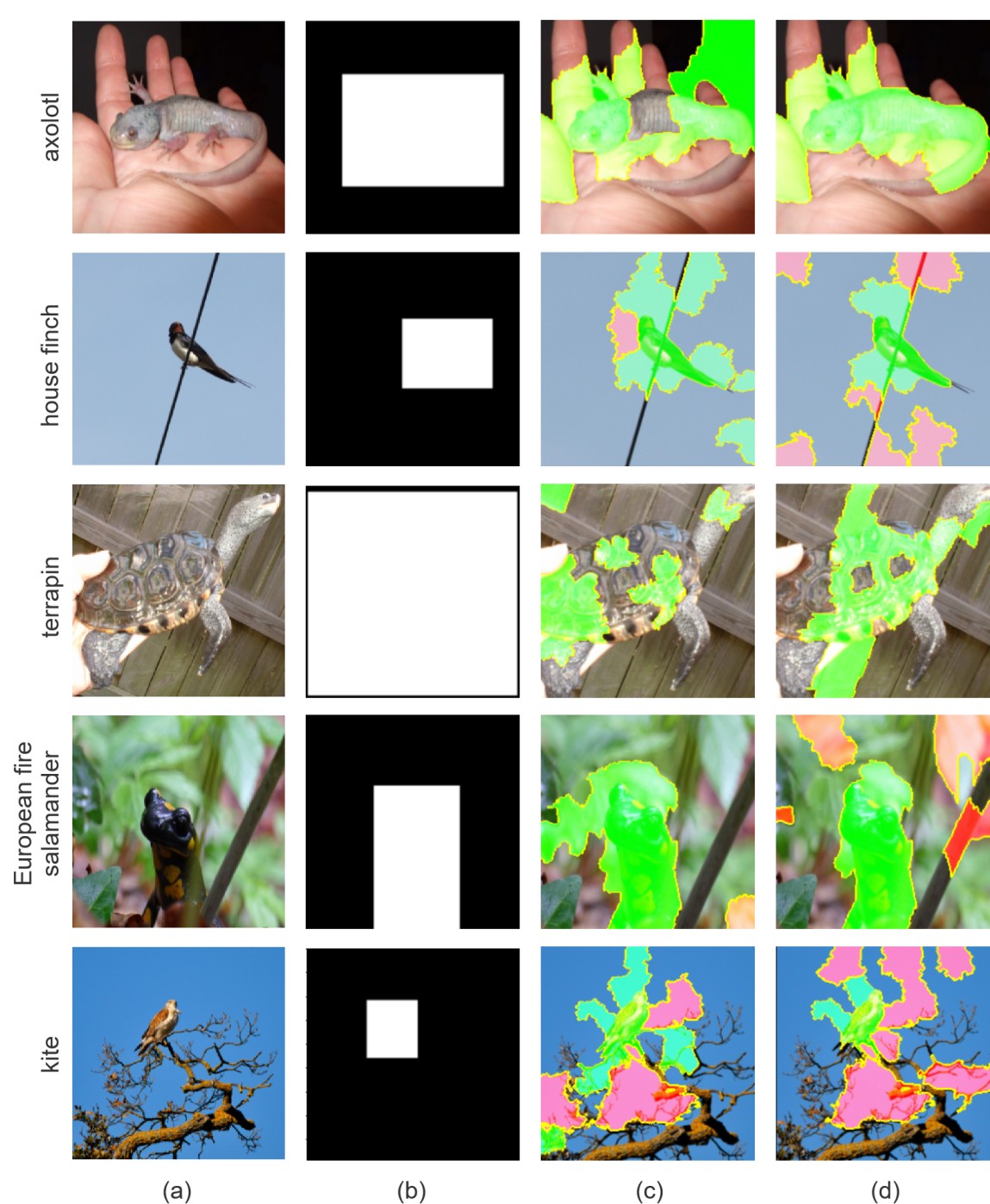}
		\caption{Evaluate LIME and $LIME^+$. (a) is the original image. (b) is the ground truth bounding box. (c) is the LIME results(composite saliency map). (d) is the $LIME^+$ results(composite saliency map).}
		\label{fig:LIME}
	\end{figure}

	LIME \cite{lime} provides green parts as positiveness to prediction and red parts as negativeness to prediction. The shade of red and green represents the weight. Unlike the two algorithms mentioned before, the results of LIME are not normal saliency maps. So it is inapplicable to use computational metrics. Here we just show some contrastive images in Fig.~\ref{fig:LIME}. $LIME^+$ labels green parts more accurately without labeling too many irrelevant areas than LIME. We cannot come to the conclusion about the red part results due to there is no ground truth on negative impact.
	
\begin{figure}[]
		\centering
		\includegraphics[scale=0.3]{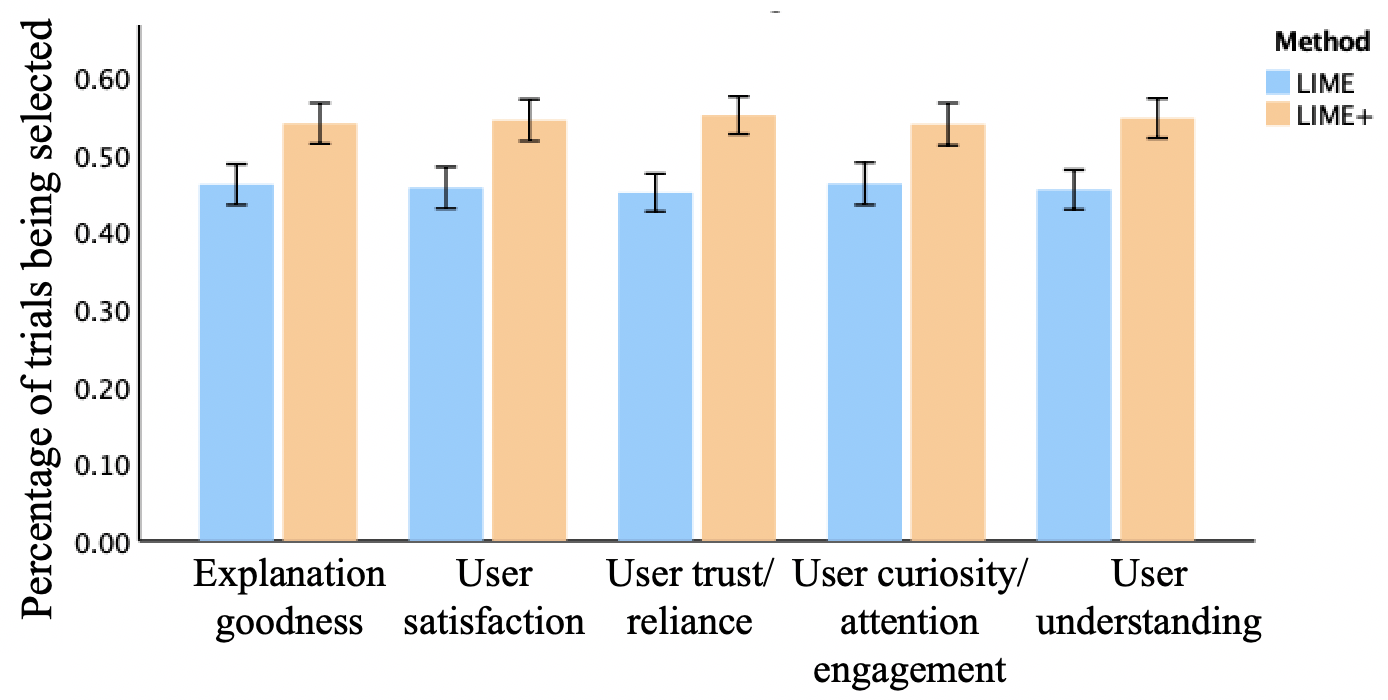}
		\caption{Percentage of trials that LIME or $LIME^+$ was selected to be a better method in terms of the 5 cognitive metrics (Error bars show one SD)}
		\label{fig:lime_cog}
	\end{figure}

	In the user study, saliency maps using $LIME^+$ were evaluated to be significantly better than those using LIME in terms of explanation goodness, t(38) = 3.04, p = .004, d = 0.49, user satisfaction, t(38) = 3.29, p = .002, d = 0.53, user curiosity/ attention engagement, t(38) = 2.84, p = .007, d = 0.45, user trust/reliance, t(38) = 4.15, p < .001, d = 0.67, and user understanding, t(38) = 3.65, p < .001, d = 0.59, as shown in Fig.~\ref{fig:lime_cog}.
	
	\subsection{Summary}
	The experiments demonstrate the advantages of our methods ($RISE^+$, $OCCLUSION^+$, $LIME^+$) in several aspects: locating the objects much more accurately, identifying more complete objects in the image, and capturing multiple objects in multi-target scenarios. In particular, $RISE^+$ provides a much more reliable saliency map for those images that RISE generates in degradation cases. The results of user study showed that our improved methods, compared with baseline methods, were consistently rated as providing better explanations, enhancing user understanding and curiosity better, and achieving higher user satisfaction and trust.

	\section{Conclusion}
	In this paper, we present an approach to improve perturbation-based XAI methods, which deals with the OoD data problem by integrating an OoD block in the generation of explanation. We discussed how to adopt this approach to three popular XAI algorithms. Our experiments show that the proposed approaches $RISE^+$, $OCCLUSION^+$, $LIME^+$ perform better than or just as well as the baseline methods based on both computational and cognitive metrics. Our solution has a wide range of applications. Other perturbation-based XAI methods with over-confidence in OoD samples is recommended to use our method. It is also suitable for various scenarios, such as the explanation of NLP, GNN models, etc.

	\bibliographystyle{IEEEtran}
	\bibliography{main}

% Generated by IEEEtran.bst, version: 1.14 (2015/08/26)
\begin{thebibliography}{10}
\providecommand{\url}[1]{#1}
\csname url@samestyle\endcsname
\providecommand{\newblock}{\relax}
\providecommand{\bibinfo}[2]{#2}
\providecommand{\BIBentrySTDinterwordspacing}{\spaceskip=0pt\relax}
\providecommand{\BIBentryALTinterwordstretchfactor}{4}
\providecommand{\BIBentryALTinterwordspacing}{\spaceskip=\fontdimen2\font plus
\BIBentryALTinterwordstretchfactor\fontdimen3\font minus
  \fontdimen4\font\relax}
\providecommand{\BIBforeignlanguage}[2]{{%
\expandafter\ifx\csname l@#1\endcsname\relax
\typeout{** WARNING: IEEEtran.bst: No hyphenation pattern has been}%
\typeout{** loaded for the language `#1'. Using the pattern for}%
\typeout{** the default language instead.}%
\else
\language=\csname l@#1\endcsname
\fi
#2}}
\providecommand{\BIBdecl}{\relax}
\BIBdecl

\bibitem{XAI_survey1}
\BIBentryALTinterwordspacing
A.~Adadi and M.~Berrada, ``Peeking inside the black-box: {A} survey on
  explainable artificial intelligence {(XAI)},'' \emph{{IEEE} Access}, vol.~6,
  pp. 52\,138--52\,160, 2018. [Online]. Available:
  \url{https://doi.org/10.1109/ACCESS.2018.2870052}
\BIBentrySTDinterwordspacing

\bibitem{XAI_survey2}
X.-H. Li, C.~C. Cao, Y.~Shi, W.~Bai, H.~Gao, L.~Qiu, C.~Wang, Y.~Gao, S.~Zhang,
  X.~Xue, and L.~Chen, ``A survey of data-driven and knowledge-aware
  explainable ai,'' \emph{IEEE Transactions on Knowledge and Data Engineering},
  pp. 1--1, 2020.

\bibitem{taxonomy1}
\BIBentryALTinterwordspacing
M.~Du, N.~Liu, and X.~Hu, ``Techniques for interpretable machine learning,''
  \emph{Commun. {ACM}}, vol.~63, no.~1, pp. 68--77, 2020. [Online]. Available:
  \url{https://doi.org/10.1145/3359786}
\BIBentrySTDinterwordspacing

\bibitem{taxonomy3}
\BIBentryALTinterwordspacing
R.~Guidotti, A.~Monreale, S.~Ruggieri, F.~Turini, F.~Giannotti, and
  D.~Pedreschi, ``A survey of methods for explaining black box models,''
  \emph{{ACM} Comput. Surv.}, vol.~51, no.~5, pp. 93:1--93:42, 2019. [Online].
  Available: \url{https://doi.org/10.1145/3236009}
\BIBentrySTDinterwordspacing

\bibitem{taxonomy4}
\BIBentryALTinterwordspacing
F.~K. Dosilovic, M.~Brcic, and N.~Hlupic, ``Explainable artificial
  intelligence: {A} survey,'' in \emph{41st International Convention on
  Information and Communication Technology, Electronics and Microelectronics,
  {MIPRO} 2018, Opatija, Croatia, May 21-25, 2018}, K.~Skala, M.~Koricic, T.~G.
  Grbac, M.~Cicin{-}Sain, V.~Sruk, S.~Ribaric, S.~Gros, B.~Vrdoljak, M.~Mauher,
  E.~Tijan, P.~Pale, and M.~Janjic, Eds.\hskip 1em plus 0.5em minus 0.4em\relax
  {IEEE}, 2018, pp. 210--215. [Online]. Available:
  \url{https://doi.org/10.23919/MIPRO.2018.8400040}
\BIBentrySTDinterwordspacing

\bibitem{taxonomy5}
O.~Biran and C.~V. Cotton, ``Explanation and justification in machine learning
  : A survey,'' 2017.

\bibitem{taxonomy6}
F.~Doshi-Velez and B.~Kim, ``Towards a rigorous science of interpretable
  machine learning,'' 2017.

\bibitem{taxonomy7}
\BIBentryALTinterwordspacing
L.~H. Gilpin, D.~Bau, B.~Z. Yuan, A.~Bajwa, M.~Specter, and L.~Kagal,
  ``Explaining explanations: An overview of interpretability of machine
  learning,'' in \emph{5th {IEEE} International Conference on Data Science and
  Advanced Analytics, {DSAA} 2018, Turin, Italy, October 1-3, 2018}, F.~Bonchi,
  F.~J. Provost, T.~Eliassi{-}Rad, W.~Wang, C.~Cattuto, and R.~Ghani,
  Eds.\hskip 1em plus 0.5em minus 0.4em\relax {IEEE}, 2018, pp. 80--89.
  [Online]. Available: \url{https://doi.org/10.1109/DSAA.2018.00018}
\BIBentrySTDinterwordspacing

\bibitem{gradcam}
\BIBentryALTinterwordspacing
R.~R. Selvaraju, M.~Cogswell, A.~Das, R.~Vedantam, D.~Parikh, and D.~Batra,
  ``Grad-cam: Visual explanations from deep networks via gradient-based
  localization,'' \emph{Int. J. Comput. Vis.}, vol. 128, no.~2, pp. 336--359,
  2020. [Online]. Available: \url{https://doi.org/10.1007/s11263-019-01228-7}
\BIBentrySTDinterwordspacing

\bibitem{aix360-sept-2019}
\BIBentryALTinterwordspacing
V.~Arya, R.~K.~E. Bellamy, P.-Y. Chen, A.~Dhurandhar, M.~Hind, S.~C. Hoffman,
  S.~Houde, Q.~V. Liao, R.~Luss, A.~Mojsilovi\'c, S.~Mourad, P.~Pedemonte,
  R.~Raghavendra, J.~Richards, P.~Sattigeri, K.~Shanmugam, M.~Singh, K.~R.
  Varshney, D.~Wei, and Y.~Zhang, ``One explanation does not fit all: A toolkit
  and taxonomy of ai explainability techniques,'' 2019. [Online]. Available:
  \url{https://arxiv.org/abs/1909.03012}
\BIBentrySTDinterwordspacing

\bibitem{captum}
\BIBentryALTinterwordspacing
Facebook, ``https://github.com/pytorch/captum.'' [Online]. Available:
  \url{https://github.com/pytorch/captum}
\BIBentrySTDinterwordspacing

\bibitem{perturbation-based_xai}
\BIBentryALTinterwordspacing
R.~C. Fong and A.~Vedaldi, ``Interpretable explanations of black boxes by
  meaningful perturbation,'' in \emph{{IEEE} International Conference on
  Computer Vision, {ICCV} 2017, Venice, Italy, October 22-29, 2017}.\hskip 1em
  plus 0.5em minus 0.4em\relax {IEEE} Computer Society, 2017, pp. 3449--3457.
  [Online]. Available: \url{https://doi.org/10.1109/ICCV.2017.371}
\BIBentrySTDinterwordspacing

\bibitem{fong2017interpretable}
\BIBentryALTinterwordspacing
S.~Kang, H.~Jung, and S.~Lee, ``Interpreting undesirable pixels for image
  classification on black-box models,'' in \emph{2019 {IEEE/CVF} International
  Conference on Computer Vision Workshops, {ICCV} Workshops 2019, Seoul, Korea
  (South), October 27-28, 2019}.\hskip 1em plus 0.5em minus 0.4em\relax {IEEE},
  2019, pp. 4250--4254. [Online]. Available:
  \url{https://doi.org/10.1109/ICCVW.2019.00523}
\BIBentrySTDinterwordspacing

\bibitem{petsiuk2018rise}
\BIBentryALTinterwordspacing
V.~Petsiuk, A.~Das, and K.~Saenko, ``{RISE:} randomized input sampling for
  explanation of black-box models,'' in \emph{British Machine Vision Conference
  2018, {BMVC} 2018, Newcastle, UK, September 3-6, 2018}.\hskip 1em plus 0.5em
  minus 0.4em\relax {BMVA} Press, 2018, p. 151. [Online]. Available:
  \url{http://bmvc2018.org/contents/papers/1064.pdf}
\BIBentrySTDinterwordspacing

\bibitem{zeiler2014visualizing}
\BIBentryALTinterwordspacing
M.~D. Zeiler and R.~Fergus, ``Visualizing and understanding convolutional
  networks,'' in \emph{Computer Vision - {ECCV} 2014 - 13th European
  Conference, Zurich, Switzerland, September 6-12, 2014, Proceedings, Part
  {I}}, ser. Lecture Notes in Computer Science, D.~J. Fleet, T.~Pajdla,
  B.~Schiele, and T.~Tuytelaars, Eds., vol. 8689.\hskip 1em plus 0.5em minus
  0.4em\relax Springer, 2014, pp. 818--833. [Online]. Available:
  \url{https://doi.org/10.1007/978-3-319-10590-1\_53}
\BIBentrySTDinterwordspacing

\bibitem{lime}
\BIBentryALTinterwordspacing
M.~T. Ribeiro, S.~Singh, and C.~Guestrin, ``"why should {I} trust you?":
  Explaining the predictions of any classifier,'' in \emph{Proceedings of the
  22nd {ACM} {SIGKDD} International Conference on Knowledge Discovery and Data
  Mining, San Francisco, CA, USA, August 13-17, 2016}, B.~Krishnapuram,
  M.~Shah, A.~J. Smola, C.~C. Aggarwal, D.~Shen, and R.~Rastogi, Eds.\hskip 1em
  plus 0.5em minus 0.4em\relax {ACM}, 2016, pp. 1135--1144. [Online].
  Available: \url{https://doi.org/10.1145/2939672.2939778}
\BIBentrySTDinterwordspacing

\bibitem{extremal}
\BIBentryALTinterwordspacing
R.~Fong, M.~Patrick, and A.~Vedaldi, ``Understanding deep networks via extremal
  perturbations and smooth masks,'' in \emph{2019 {IEEE/CVF} International
  Conference on Computer Vision, {ICCV} 2019, Seoul, Korea (South), October 27
  - November 2, 2019}.\hskip 1em plus 0.5em minus 0.4em\relax {IEEE}, 2019, pp.
  2950--2958. [Online]. Available:
  \url{https://doi.org/10.1109/ICCV.2019.00304}
\BIBentrySTDinterwordspacing

\bibitem{OOD_inpainting}
\BIBentryALTinterwordspacing
C.~Agarwal and A.~Nguyen, ``Explaining image classifiers by removing input
  features using generative models,'' in \emph{Computer Vision - {ACCV} 2020 -
  15th Asian Conference on Computer Vision, Kyoto, Japan, November 30 -
  December 4, 2020, Revised Selected Papers, Part {VI}}, ser. Lecture Notes in
  Computer Science, H.~Ishikawa, C.~Liu, T.~Pajdla, and J.~Shi, Eds., vol.
  12627.\hskip 1em plus 0.5em minus 0.4em\relax Springer, 2020, pp. 101--118.
  [Online]. Available: \url{https://doi.org/10.1007/978-3-030-69544-6\_7}
\BIBentrySTDinterwordspacing

\bibitem{nguyen2015deep}
\BIBentryALTinterwordspacing
A.~M. Nguyen, J.~Yosinski, and J.~Clune, ``Deep neural networks are easily
  fooled: High confidence predictions for unrecognizable images,'' in
  \emph{{IEEE} Conference on Computer Vision and Pattern Recognition, {CVPR}
  2015, Boston, MA, USA, June 7-12, 2015}.\hskip 1em plus 0.5em minus
  0.4em\relax {IEEE} Computer Society, 2015, pp. 427--436. [Online]. Available:
  \url{https://doi.org/10.1109/CVPR.2015.7298640}
\BIBentrySTDinterwordspacing

\bibitem{OoD_detect_baseline}
\BIBentryALTinterwordspacing
D.~Hendrycks and K.~Gimpel, ``A baseline for detecting misclassified and
  out-of-distribution examples in neural networks,'' in \emph{5th International
  Conference on Learning Representations, {ICLR} 2017, Toulon, France, April
  24-26, 2017, Conference Track Proceedings}.\hskip 1em plus 0.5em minus
  0.4em\relax OpenReview.net, 2017. [Online]. Available:
  \url{https://openreview.net/forum?id=Hkg4TI9xl}
\BIBentrySTDinterwordspacing

\bibitem{szegedy2014intriguing}
\BIBentryALTinterwordspacing
C.~Szegedy, W.~Zaremba, I.~Sutskever, J.~Bruna, D.~Erhan, I.~J. Goodfellow, and
  R.~Fergus, ``Intriguing properties of neural networks,'' in \emph{2nd
  International Conference on Learning Representations, {ICLR} 2014, Banff, AB,
  Canada, April 14-16, 2014, Conference Track Proceedings}, Y.~Bengio and
  Y.~LeCun, Eds., 2014. [Online]. Available:
  \url{http://arxiv.org/abs/1312.6199}
\BIBentrySTDinterwordspacing

\bibitem{chang2018explaining}
\BIBentryALTinterwordspacing
C.~Chang, E.~Creager, A.~Goldenberg, and D.~Duvenaud, ``Explaining image
  classifiers by counterfactual generation,'' in \emph{7th International
  Conference on Learning Representations, {ICLR} 2019, New Orleans, LA, USA,
  May 6-9, 2019}.\hskip 1em plus 0.5em minus 0.4em\relax OpenReview.net, 2019.
  [Online]. Available: \url{https://openreview.net/forum?id=B1MXz20cYQ}
\BIBentrySTDinterwordspacing

\bibitem{simonyan2013deep}
\BIBentryALTinterwordspacing
K.~Simonyan, A.~Vedaldi, and A.~Zisserman, ``Deep inside convolutional
  networks: Visualising image classification models and saliency maps,'' in
  \emph{2nd International Conference on Learning Representations, {ICLR} 2014,
  Banff, AB, Canada, April 14-16, 2014, Workshop Track Proceedings}, Y.~Bengio
  and Y.~LeCun, Eds., 2014. [Online]. Available:
  \url{http://arxiv.org/abs/1312.6034}
\BIBentrySTDinterwordspacing

\bibitem{rudin2019stop}
C.~Rudin, ``Stop explaining black box machine learning models for high stakes
  decisions and use interpretable models instead,'' \emph{Nature Machine
  Intelligence}, vol.~1, no.~5, pp. 206--215, 2019.

\bibitem{ghorbani2019interpretation}
\BIBentryALTinterwordspacing
A.~Ghorbani, A.~Abid, and J.~Y. Zou, ``Interpretation of neural networks is
  fragile,'' in \emph{The Thirty-Third {AAAI} Conference on Artificial
  Intelligence, {AAAI} 2019, The Thirty-First Innovative Applications of
  Artificial Intelligence Conference, {IAAI} 2019, The Ninth {AAAI} Symposium
  on Educational Advances in Artificial Intelligence, {EAAI} 2019, Honolulu,
  Hawaii, USA, January 27 - February 1, 2019}.\hskip 1em plus 0.5em minus
  0.4em\relax {AAAI} Press, 2019, pp. 3681--3688. [Online]. Available:
  \url{https://doi.org/10.1609/aaai.v33i01.33013681}
\BIBentrySTDinterwordspacing

\bibitem{slack2020fooling}
\BIBentryALTinterwordspacing
D.~Slack, S.~Hilgard, E.~Jia, S.~Singh, and H.~Lakkaraju, ``Fooling {LIME} and
  {SHAP:} adversarial attacks on post hoc explanation methods,'' in
  \emph{{AIES} '20: {AAAI/ACM} Conference on AI, Ethics, and Society, New York,
  NY, USA, February 7-8, 2020}, A.~N. Markham, J.~Powles, T.~Walsh, and A.~L.
  Washington, Eds.\hskip 1em plus 0.5em minus 0.4em\relax {ACM}, 2020, pp.
  180--186. [Online]. Available: \url{https://doi.org/10.1145/3375627.3375830}
\BIBentrySTDinterwordspacing

\bibitem{SHAP}
\BIBentryALTinterwordspacing
S.~M. Lundberg and S.~Lee, ``A unified approach to interpreting model
  predictions,'' in \emph{Advances in Neural Information Processing Systems 30:
  Annual Conference on Neural Information Processing Systems 2017, December
  4-9, 2017, Long Beach, CA, {USA}}, I.~Guyon, U.~von Luxburg, S.~Bengio, H.~M.
  Wallach, R.~Fergus, S.~V.~N. Vishwanathan, and R.~Garnett, Eds., 2017, pp.
  4765--4774. [Online]. Available:
  \url{https://proceedings.neurips.cc/paper/2017/hash/8a20a8621978632d76c43dfd28b67767-Abstract.html}
\BIBentrySTDinterwordspacing

\bibitem{AnomalyDetection1}
\BIBentryALTinterwordspacing
P.~Oberdiek, M.~Rottmann, and H.~Gottschalk, ``Classification uncertainty of
  deep neural networks based on gradient information,'' in \emph{Artificial
  Neural Networks in Pattern Recognition - 8th {IAPR} {TC3} Workshop, {ANNPR}
  2018, Siena, Italy, September 19-21, 2018, Proceedings}, ser. Lecture Notes
  in Computer Science, L.~Pancioni, F.~Schwenker, and E.~Trentin, Eds., vol.
  11081.\hskip 1em plus 0.5em minus 0.4em\relax Springer, 2018, pp. 113--125.
  [Online]. Available: \url{https://doi.org/10.1007/978-3-319-99978-4\_9}
\BIBentrySTDinterwordspacing

\bibitem{AnomalyDetection2}
\BIBentryALTinterwordspacing
H.~Jiang, B.~Kim, M.~Y. Guan, and M.~R. Gupta, ``To trust or not to trust {A}
  classifier,'' in \emph{Advances in Neural Information Processing Systems 31:
  Annual Conference on Neural Information Processing Systems 2018, NeurIPS
  2018, December 3-8, 2018, Montr{\'{e}}al, Canada}, S.~Bengio, H.~M. Wallach,
  H.~Larochelle, K.~Grauman, N.~Cesa{-}Bianchi, and R.~Garnett, Eds., 2018, pp.
  5546--5557. [Online]. Available:
  \url{https://proceedings.neurips.cc/paper/2018/hash/7180cffd6a8e829dacfc2a31b3f72ece-Abstract.html}
\BIBentrySTDinterwordspacing

\bibitem{AnomalyDetection3}
\BIBentryALTinterwordspacing
S.~Liu, R.~Garrepalli, T.~G. Dietterich, A.~Fern, and D.~Hendrycks, ``Open
  category detection with {PAC} guarantees,'' in \emph{Proceedings of the 35th
  International Conference on Machine Learning, {ICML} 2018,
  Stockholmsm{\"{a}}ssan, Stockholm, Sweden, July 10-15, 2018}, ser.
  Proceedings of Machine Learning Research, J.~G. Dy and A.~Krause, Eds.,
  vol.~80.\hskip 1em plus 0.5em minus 0.4em\relax {PMLR}, 2018, pp. 3175--3184.
  [Online]. Available: \url{http://proceedings.mlr.press/v80/liu18e.html}
\BIBentrySTDinterwordspacing

\bibitem{AnomalyDetection4}
\BIBentryALTinterwordspacing
J.~Ren, P.~J. Liu, E.~Fertig, J.~Snoek, R.~Poplin, M.~A. DePristo, J.~V.
  Dillon, and B.~Lakshminarayanan, ``Likelihood ratios for out-of-distribution
  detection,'' in \emph{Advances in Neural Information Processing Systems 32:
  Annual Conference on Neural Information Processing Systems 2019, NeurIPS
  2019, December 8-14, 2019, Vancouver, BC, Canada}, H.~M. Wallach,
  H.~Larochelle, A.~Beygelzimer, F.~d'Alch{\'{e}}{-}Buc, E.~B. Fox, and
  R.~Garnett, Eds., 2019, pp. 14\,680--14\,691. [Online]. Available:
  \url{https://proceedings.neurips.cc/paper/2019/hash/1e79596878b2320cac26dd792a6c51c9-Abstract.html}
\BIBentrySTDinterwordspacing

\bibitem{AnomalyDetection5}
\BIBentryALTinterwordspacing
Q.~Yu and K.~Aizawa, ``Unsupervised out-of-distribution detection by maximum
  classifier discrepancy,'' in \emph{2019 {IEEE/CVF} International Conference
  on Computer Vision, {ICCV} 2019, Seoul, Korea (South), October 27 - November
  2, 2019}.\hskip 1em plus 0.5em minus 0.4em\relax {IEEE}, 2019, pp.
  9517--9525. [Online]. Available:
  \url{https://doi.org/10.1109/ICCV.2019.00961}
\BIBentrySTDinterwordspacing

\bibitem{AnomalyDetection7}
\BIBentryALTinterwordspacing
S.~Liang, Y.~Li, and R.~Srikant, ``Principled detection of out-of-distribution
  examples in neural networks,'' \emph{CoRR}, vol. abs/1706.02690, 2017.
  [Online]. Available: \url{http://arxiv.org/abs/1706.02690}
\BIBentrySTDinterwordspacing

\bibitem{AnomalyDetection8}
\BIBentryALTinterwordspacing
W.~E. Lawson, E.~Bekele, and K.~Sullivan, ``Finding anomalies with generative
  adversarial networks for a patrolbot,'' in \emph{2017 {IEEE} Conference on
  Computer Vision and Pattern Recognition Workshops, {CVPR} Workshops 2017,
  Honolulu, HI, USA, July 21-26, 2017}.\hskip 1em plus 0.5em minus 0.4em\relax
  {IEEE} Computer Society, 2017, pp. 484--485. [Online]. Available:
  \url{https://doi.org/10.1109/CVPRW.2017.68}
\BIBentrySTDinterwordspacing

\bibitem{AnomalyDetection9}
\BIBentryALTinterwordspacing
W.~Chen, Y.~Shen, X.~Wang, and W.~Y. Wang, ``Enhancing the robustness of prior
  network in out-of-distribution detection,'' \emph{CoRR}, vol. abs/1811.07308,
  2018. [Online]. Available: \url{http://arxiv.org/abs/1811.07308}
\BIBentrySTDinterwordspacing

\bibitem{AnomalyDetection10}
\BIBentryALTinterwordspacing
I.~Golan and R.~El{-}Yaniv, ``Deep anomaly detection using geometric
  transformations,'' in \emph{Advances in Neural Information Processing Systems
  31: Annual Conference on Neural Information Processing Systems 2018, NeurIPS
  2018, December 3-8, 2018, Montr{\'{e}}al, Canada}, S.~Bengio, H.~M. Wallach,
  H.~Larochelle, K.~Grauman, N.~Cesa{-}Bianchi, and R.~Garnett, Eds., 2018, pp.
  9781--9791. [Online]. Available:
  \url{https://proceedings.neurips.cc/paper/2018/hash/5e62d03aec0d17facfc5355dd90d441c-Abstract.html}
\BIBentrySTDinterwordspacing

\bibitem{AnomalyDetection11}
\BIBentryALTinterwordspacing
T.~Denouden, R.~Salay, K.~Czarnecki, V.~Abdelzad, B.~Phan, and S.~Vernekar,
  ``Improving reconstruction autoencoder out-of-distribution detection with
  mahalanobis distance,'' \emph{CoRR}, vol. abs/1812.02765, 2018. [Online].
  Available: \url{http://arxiv.org/abs/1812.02765}
\BIBentrySTDinterwordspacing

\bibitem{AnomalyDetection12}
\BIBentryALTinterwordspacing
A.~Shafaei, M.~Schmidt, and J.~J. Little, ``Does your model know the digit 6 is
  not a cat? {A} less biased evaluation of "outlier" detectors,'' \emph{CoRR},
  vol. abs/1809.04729, 2018. [Online]. Available:
  \url{http://arxiv.org/abs/1809.04729}
\BIBentrySTDinterwordspacing

\bibitem{GODIN}
\BIBentryALTinterwordspacing
Y.~Hsu, Y.~Shen, H.~Jin, and Z.~Kira, ``Generalized {ODIN:} detecting
  out-of-distribution image without learning from out-of-distribution data,''
  in \emph{2020 {IEEE/CVF} Conference on Computer Vision and Pattern
  Recognition, {CVPR} 2020, Seattle, WA, USA, June 13-19, 2020}.\hskip 1em plus
  0.5em minus 0.4em\relax {IEEE}, 2020, pp. 10\,948--10\,957. [Online].
  Available: \url{https://doi.org/10.1109/CVPR42600.2020.01096}
\BIBentrySTDinterwordspacing

\bibitem{metric}
\BIBentryALTinterwordspacing
X.~Li, Y.~Shi, H.~Li, W.~Bai, Y.~Song, C.~C. Cao, and L.~Chen, ``Quantitative
  evaluations on saliency methods: An experimental study,'' \emph{CoRR}, vol.
  abs/2012.15616, 2020. [Online]. Available:
  \url{https://arxiv.org/abs/2012.15616}
\BIBentrySTDinterwordspacing

\bibitem{zhou2016learning}
\BIBentryALTinterwordspacing
B.~Zhou, A.~Khosla, {\`{A}}.~Lapedriza, A.~Oliva, and A.~Torralba, ``Learning
  deep features for discriminative localization,'' in \emph{2016 {IEEE}
  Conference on Computer Vision and Pattern Recognition, {CVPR} 2016, Las
  Vegas, NV, USA, June 27-30, 2016}.\hskip 1em plus 0.5em minus 0.4em\relax
  {IEEE} Computer Society, 2016, pp. 2921--2929. [Online]. Available:
  \url{https://doi.org/10.1109/CVPR.2016.319}
\BIBentrySTDinterwordspacing

\bibitem{zhang2018top}
\BIBentryALTinterwordspacing
J.~Zhang, Z.~L. Lin, J.~Brandt, X.~Shen, and S.~Sclaroff, ``Top-down neural
  attention by excitation backprop,'' in \emph{Computer Vision - {ECCV} 2016 -
  14th European Conference, Amsterdam, The Netherlands, October 11-14, 2016,
  Proceedings, Part {IV}}, ser. Lecture Notes in Computer Science, B.~Leibe,
  J.~Matas, N.~Sebe, and M.~Welling, Eds., vol. 9908.\hskip 1em plus 0.5em
  minus 0.4em\relax Springer, 2016, pp. 543--559. [Online]. Available:
  \url{https://doi.org/10.1007/978-3-319-46493-0\_33}
\BIBentrySTDinterwordspacing

\bibitem{wang2019designing}
D.~Wang, Q.~Yang, A.~Abdul, and B.~Y. Lim, ``Designing theory-driven
  user-centric explainable ai,'' in \emph{Proceedings of the 2019 CHI
  conference on human factors in computing systems}, 2019, pp. 1--15.

\bibitem{Hoffman2018}
\BIBentryALTinterwordspacing
R.~R. Hoffman, S.~T. Mueller, G.~Klein, and J.~Litman, ``Metrics for
  explainable {AI:} challenges and prospects,'' \emph{CoRR}, vol.
  abs/1812.04608, 2018. [Online]. Available:
  \url{http://arxiv.org/abs/1812.04608}
\BIBentrySTDinterwordspacing

\bibitem{jin2021euca}
W.~Jin, J.~Fan, D.~Gromala, P.~Pasquier, and G.~Hamarneh, ``Euca: A practical
  prototyping framework towards end-user-centered explainable artificial
  intelligence,'' \emph{arXiv preprint arXiv:2102.02437}, 2021.

\bibitem{holzinger2020measuring}
A.~Holzinger, A.~Carrington, and H.~M{\"u}ller, ``Measuring the quality of
  explanations: the system causability scale (scs),'' \emph{KI-K{\"u}nstliche
  Intelligenz}, pp. 1--6, 2020.

\bibitem{kim2016examples}
B.~Kim, R.~Khanna, and O.~O. Koyejo, ``Examples are not enough, learn to
  criticize! criticism for interpretability,'' \emph{Advances in neural
  information processing systems}, vol.~29, 2016.

\bibitem{cohen1992statistical}
J.~Cohen, ``Statistical power analysis,'' \emph{Current directions in
  psychological science}, vol.~1, no.~3, pp. 98--101, 1992.

\bibitem{ODIN}
\BIBentryALTinterwordspacing
S.~Liang, Y.~Li, and R.~Srikant, ``Enhancing the reliability of
  out-of-distribution image detection in neural networks,'' in \emph{6th
  International Conference on Learning Representations, {ICLR} 2018, Vancouver,
  BC, Canada, April 30 - May 3, 2018, Conference Track Proceedings}.\hskip 1em
  plus 0.5em minus 0.4em\relax OpenReview.net, 2018. [Online]. Available:
  \url{https://openreview.net/forum?id=H1VGkIxRZ}
\BIBentrySTDinterwordspacing

\bibitem{preprocessing}
\BIBentryALTinterwordspacing
I.~J. Goodfellow, J.~Shlens, and C.~Szegedy, ``Explaining and harnessing
  adversarial examples,'' in \emph{3rd International Conference on Learning
  Representations, {ICLR} 2015, San Diego, CA, USA, May 7-9, 2015, Conference
  Track Proceedings}, Y.~Bengio and Y.~LeCun, Eds., 2015. [Online]. Available:
  \url{http://arxiv.org/abs/1412.6572}
\BIBentrySTDinterwordspacing

\bibitem{highdimensional_curse}
A.~Koufakou and M.~Georgiopoulos, ``A fast outlier detection strategy for
  distributed high-dimensional data sets with mixed attributes,'' \emph{Data
  Min. Knowl. Discov.}, vol.~20, pp. 259--289, 03 2010.

\bibitem{NEURIPS2018_faithfulness}
\BIBentryALTinterwordspacing
D.~Alvarez{-}Melis and T.~S. Jaakkola, ``Towards robust interpretability with
  self-explaining neural networks,'' in \emph{Advances in Neural Information
  Processing Systems 31: Annual Conference on Neural Information Processing
  Systems 2018, NeurIPS 2018, December 3-8, 2018, Montr{\'{e}}al, Canada},
  S.~Bengio, H.~M. Wallach, H.~Larochelle, K.~Grauman, N.~Cesa{-}Bianchi, and
  R.~Garnett, Eds., 2018, pp. 7786--7795. [Online]. Available:
  \url{https://proceedings.neurips.cc/paper/2018/hash/3e9f0fc9b2f89e043bc6233994dfcf76-Abstract.html}
\BIBentrySTDinterwordspacing

\bibitem{MILLER20191}
\BIBentryALTinterwordspacing
T.~Miller, ``Explanation in artificial intelligence: Insights from the social
  sciences,'' \emph{Artificial Intelligence}, vol. 267, pp. 1--38, 2019.
  [Online]. Available:
  \url{https://www.sciencedirect.com/science/article/pii/S0004370218305988}
\BIBentrySTDinterwordspacing

\bibitem{Hsiao2021}
J.~HSIAO, H.~H.~T. NGAI, L.~QIU, Y.~YANG, and C.~C. CAO, ``Roadmap for
  designing cognitive metrics for explainable artificial intelligence (xai).''
  \emph{Artificial Intelligence}, 2021.

\bibitem{he2016deep}
\BIBentryALTinterwordspacing
K.~He, X.~Zhang, S.~Ren, and J.~Sun, ``Deep residual learning for image
  recognition,'' in \emph{2016 {IEEE} Conference on Computer Vision and Pattern
  Recognition, {CVPR} 2016, Las Vegas, NV, USA, June 27-30, 2016}.\hskip 1em
  plus 0.5em minus 0.4em\relax {IEEE} Computer Society, 2016, pp. 770--778.
  [Online]. Available: \url{https://doi.org/10.1109/CVPR.2016.90}
\BIBentrySTDinterwordspacing

\bibitem{vgg16}
\BIBentryALTinterwordspacing
K.~Simonyan and A.~Zisserman, ``Very deep convolutional networks for
  large-scale image recognition,'' in \emph{3rd International Conference on
  Learning Representations, {ICLR} 2015, San Diego, CA, USA, May 7-9, 2015,
  Conference Track Proceedings}, Y.~Bengio and Y.~LeCun, Eds., 2015. [Online].
  Available: \url{http://arxiv.org/abs/1409.1556}
\BIBentrySTDinterwordspacing

\bibitem{russakovsky2015imagenet}
\BIBentryALTinterwordspacing
O.~Russakovsky, J.~Deng, H.~Su, J.~Krause, S.~Satheesh, S.~Ma, Z.~Huang,
  A.~Karpathy, A.~Khosla, M.~S. Bernstein, A.~C. Berg, and F.~Li, ``Imagenet
  large scale visual recognition challenge,'' \emph{Int. J. Comput. Vis.}, vol.
  115, no.~3, pp. 211--252, 2015. [Online]. Available:
  \url{https://doi.org/10.1007/s11263-015-0816-y}
\BIBentrySTDinterwordspacing

\bibitem{voc}
M.~Everingham, L.~Van~Gool, C.~K.~I. Williams, J.~Winn, and A.~Zisserman, ``The
  {PASCAL} {V}isual {O}bject {C}lasses {C}hallenge 2007 {(VOC2007)}
  {R}esults,''
  http://www.pascal-network.org/challenges/VOC/voc2007/workshop/index.html.

\bibitem{cahour2009d}
B.~Cahour and J.-F. Forzy, ``Does projection into use improve trust and
  exploration? an example with a cruise control system,'' \emph{Safety
  science}, vol.~47, no.~9, pp. 1260--1270, 2009.

\bibitem{jian2000foundations}
J.-Y. Jian, A.~M. Bisantz, and C.~G. Drury, ``Foundations for an empirically
  determined scale of trust in automated systems,'' \emph{International journal
  of cognitive ergonomics}, vol.~4, no.~1, pp. 53--71, 2000.

\bibitem{schaefer2013perception}
K.~Schaefer, ``The perception and measurement of human-robot trust,'' 2013.

\bibitem{madsen2000measuring}
M.~Madsen and S.~Gregor, ``Measuring human-computer trust,'' in \emph{11th
  australasian conference on information systems}, vol.~53.\hskip 1em plus
  0.5em minus 0.4em\relax Citeseer, 2000, pp. 6--8.

\end{thebibliography}
	
\end{document}